\documentclass[lettersize,journal]{IEEEtran}
\usepackage{amsmath,amsfonts}
\usepackage{algorithmic}
\usepackage{algorithm}
\usepackage{array}
\usepackage[caption=false,font=normalsize,labelfont=sf,textfont=sf]{subfig}
\usepackage{textcomp}
\usepackage{stfloats}
\usepackage{url}
\usepackage{verbatim}
\usepackage{graphicx}
\usepackage{cite}
\usepackage{booktabs}
\usepackage{multirow}
\graphicspath{{./figs/}}
\begin{document}

\title{Breakout-picker: Reducing false positives in deep learning-based borehole breakout characterization from acoustic image logs}

\author{Guangyu Wang, Xiaodong Ma, and Xinming Wu,~\IEEEmembership{Associate member,~IEEE}
\thanks{The authors are with the State Key Laboratory of Precision Geodesy,
School of Earth and Space Sciences, University of Science and Technology
of China, Hefei 230026, China, also with Mengcheng National Geophysical
Observatory, University of Science and Technology of China, Mengcheng
233500, China}
}



\maketitle

\begin{abstract}
Borehole breakouts are stress-induced spalling on the borehole wall, which are identifiable in acoustic image logs as paired zones with near-symmetry azimuths, low acoustic amplitudes, and increased borehole radius. Accurate breakout characterization is crucial for in-situ stress analysis. In recent years, deep learning has been introduced to automate the time-consuming and labor-intensive breakout picking process. However, existing approaches often suffer from misclassification of non-breakout features, leading to high false positive rates. To address this limitation, this study develops a deep learning framework, termed Breakout-picker, with a specific focus on reducing false positives in automatic breakout characterization. Breakout-picker reduces false positives through two strategies. First, the training of Breakout-picker incorporates negative samples of non-breakout features, including natural fractures, keyseats, and logging artifacts. They share similar characteristics with breakouts, such as low acoustic amplitude or locally enlarged borehole radius. These negative training samples enables Breakout-picker to better discriminate true breakouts and similar non-breakout features. Second, candidate breakouts identified by Breakout-picker are further validated by azimuthal symmetry criteria, whereby detections that do not exhibit the near-symmetry characteristics of breakout azimuth are excluded. The performance of Breakout-picker is evaluated using three acoustic image log datasets from different regions. The results demonstrate that Breakout-picker outperforms other automatic methods with higher accuracy and substantially lower false positive rates. By reducing false positives, Breakout-picker enhances the reliability of automatic breakout characterization from acoustic image logs, which in turn benefits in-situ stress analysis based on borehole breakouts.
\end{abstract}


\section{Introduction}
Borehole breakouts are spalled regions of boreholes due to stress-induced compressive failure of the surrounding rock. They typically appear as symmetrical zones of borehole wall damage, aligning with the direction of minimum horizontal stress \cite{bell1979northeast, blumling1983orientation, zoback1985well}. In geoscientific investigation and subsurface resource development, borehole breakouts have been widely used as a robust indicator for characterizing in-situ stress (e.g., \cite{moos1990utilization, zoback2003determination, lin2010localized, valley2019stress}). The azimuth of breakout reflects stress orientations, and the geometry of breakout can be used to constrain rock strength and stress magnitudes.

Borehole breakouts can be identified using various downhole logs, including petrophysical logs such as density, gamma ray, and resistivity logs (e.g., \cite{bashmagh2024application}), as well as caliper logs (e.g., \cite{plumb1985stress}), electrical image logs (e.g., \cite{rajabi2010subsurface}), and acoustic image logs (e.g., \cite{zoback2003determination}). Petrophysical logs can indirectly indicate the presence of breakouts through anomalous responses in formation parameters, but they do not provide information on breakout azimuths or geometries. Caliper logs detect borehole enlargements and can estimate breakout azimuths, yet they offer limited constraints on breakout geometry. Electrical image logs allow direct imaging of the borehole wall. However, breakout-related borehole spalling and gaps between tool pads often result in poorly resolved or incomplete breakout images. By contrast, acoustic image logging, which emits ultrasonic waves towards the borehole wall and records the reflected signals, provide full-bore coverage of the borehole wall and are well suited for characterizing both breakout azimuths and geometries. An acoustic image log consists of two components: the amplitudes and travel-times of the reflected acoustic waves, where the travel-times can be used to derive borehole radius. Breakouts in acoustic image logs typically appear as broad zones of enlarged borehole radius and reduced acoustic amplitude, making acoustic imaging particularly effective for breakout characterization.

Breakout characterization involves determining the depths, azimuths, and widths of borehole breakouts and is traditionally performed manually by experienced experts. This manual workflow is time-consuming and labor-intensive, particularly for boreholes exhibiting long and continuous breakout zones. With the rapidly increasing volume of acoustic image logs, automatic breakout characterization has therefore become essential for efficient and reliable in-situ stress analysis. In response, two main classes of automatic methods have been developed. The first class is based on peak-detection approaches that exploit the physical characteristic of breakouts in acoustic image logs. In this method, breakout azimuths are determined by locating minimum acoustic amplitude and maximum borehole radius, while breakout widths are estimated from the azimuthal separation between the detected breakout edges (e.g., \cite{wessling2011automated, wang2020automated}). The second class of methods relies on deep learning (DL). Convolutional neural networks (CNNs) have been trained for breakout detection (e.g., \cite{dias2020automatic, yeom2025automatic}) as well as for breakout segmentation (e.g., \cite{yang2022automated, cunha2025semi}). Breakout detection produces bounding boxes that localize breakout regions, whereas breakout segmentation yields pixel-level masks that delineate breakout geometries in greater detail. Both approaches enable the subsequent extraction of breakout azimuths and widths.

Despite their demonstrative effectiveness, both classes of automatic breakout characterization methods have inherent limitations. The peak-detection approach relies on predefined signal patterns in acoustic image logs to identify breakouts. As a result, they are sensitive to logging artifacts and borehole wall irregularities, which can produce similar signal patterns and lead to misidentification. DL-based methods, on the other hand, automatically learn breakout features from labeled training data without explicit rules, making them more adaptable to variations in borehole conditions and logging responses. However, features with appearance similar to breakouts, such as keyseats or other borehole enlargements, may be misidentify as breakouts, resulting in false positive predictions. These false positives can degrade the reliability of derived breakout parameters (i.e., depths, azimuths, and widths) and subsequent in-situ stress analysis. Therefore, reducing false positives remains a key challenge for improving the robustness of DL-based automatic breakout characterization.

In this study, we propose a DL-based framework, termed Breakout-picker, for reducing false positives in breakout characterization from acoustic image logs. Breakout-picker employs the DeepLabV3+ model \cite{chen2018encoder} to segment breakout regions and subsequently extracts breakout parameters from the segmentation results. False positives are reduced through two strategies. First, whereas previous DL-based approaches typically rely solely on positive training samples (i.e., log segments containing breakouts), we adopt a training strategy that incorporates both positive samples and negative samples. These negative samples consist of log segments without breakouts but containing features that are similar to breakouts, such as keyseats and large-opening fractures. Including such negative samples is critical for enabling DL models to distinguish true breakouts from non-breakout features with similar appearances. Second, a post-processing step is applied to further suppress false positives. Candidate breakouts identified by the DeepLabV3+ model are evaluated using azimuthal symmetry criteria, and detections that do not satisfy the expected near-symmetric azimuthal characteristics of breakouts are excluded. Applications to field acoustic image logs demonstrate that Breakout-picker outperforms other automatic approaches, yielding accurate breakout characterization with substantially fewer false positives.

\section{Method}
\subsection{Breakout-picker framework}
\begin{figure*}[t]
\centering
\includegraphics[width=0.8\textwidth]{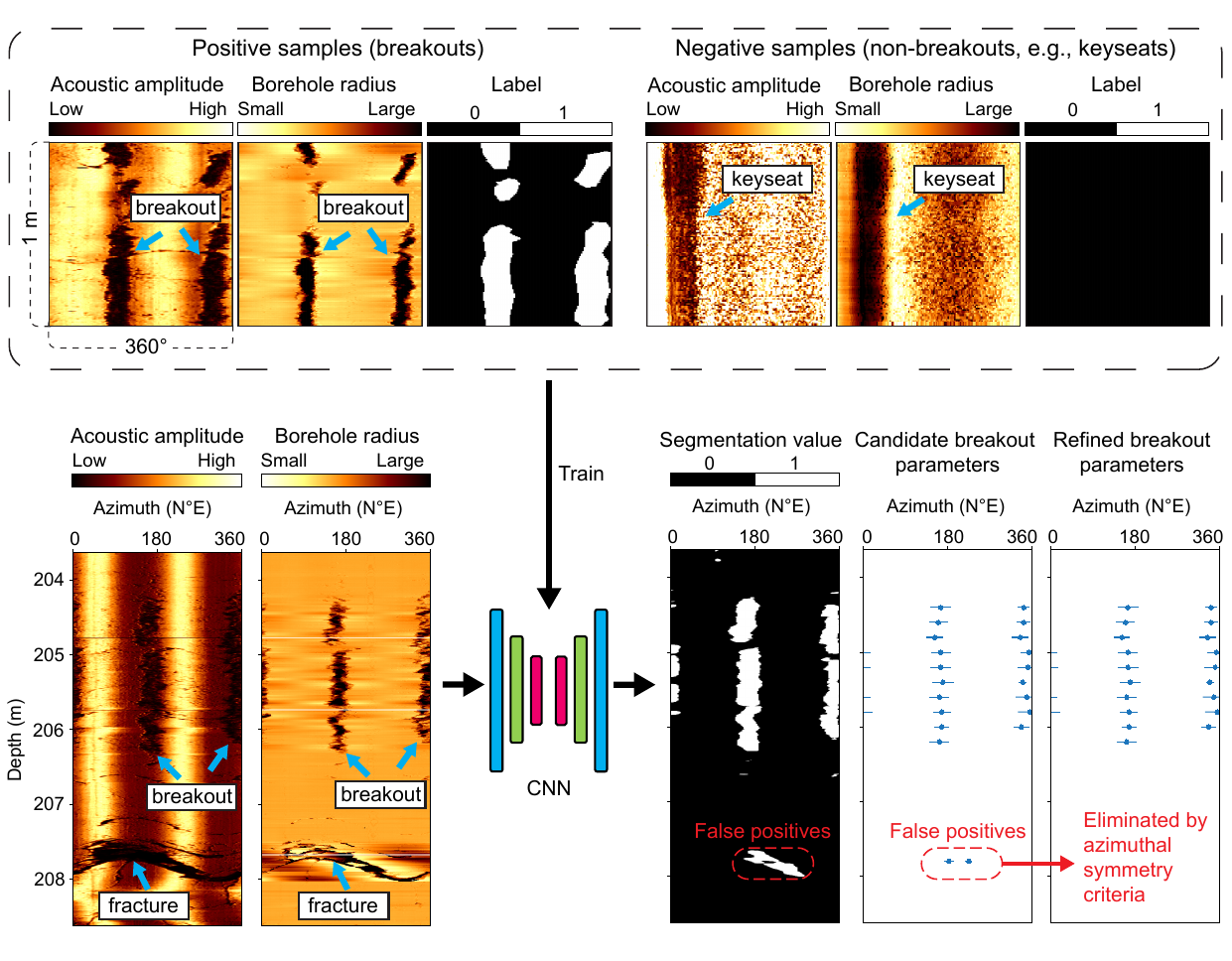}
\caption{Overview of the Breakout-picker framework for automatic borehole breakout characterization with reduced false positives. In the label/segmentation images, the black (0) and white (1) regions denote non-breakout and breakout areas, respectively. The breakout parameters are visualized using blue error bars, with points denoting breakout azimuths and bars denoting the azimuthal span (i.e., widths) of the breakouts.}
\label{fig1}
\end{figure*}

The framework is designed to combine the pattern recognition capability of deep learning with physically motivated breakout characteristics, aiming to achieve reliable automatic borehole breakout characterization with reduced false positives. An overview of the Breakout-picker framework is illustrated in Figure \ref{fig1}. In this framework, a CNN is used to identify candidate breakout regions based on acoustic image logs, while the final breakout determination is constrained by an azimuthal symmetry criterion derived from borehole breakout mechanics.

\subsubsection{Breakout segmentation based on CNN}
The CNN is trained using segments of acoustic amplitude logs and borehole radius logs as dual-channel inputs, together with manually annotated breakout labels. The CNN performs pixel-wise segmentation and outputs a binary image, where breakout regions are denoted by ones and the background by zeros. Pixel-wise segmentation is adopted instead of direct azimuth prediction because it preserves the spatial extent and geometry of breakout regions, which are essential for subsequent extraction of breakout parameters and physically motivated validation. The CNN output is therefore treated as candidate breakout segmentation rather than final breakout identification.

Both positive and negative samples are included during training. Positive samples correspond to true borehole breakouts. Negative samples are manually selected non-breakout intervals that exhibit visual characteristics similar to breakouts, such as low acoustic amplitude or locally enlarged borehole radius. These negative samples include features such as keyseats, large-opening fractures, and logging-related artifacts (e.g., the keyseat example shown in Figure \ref{fig1}). Such non-breakout features represent the primary source of false positives in automatic breakout characterization. Rather than using randomly selected background samples, these hard negative samples are deliberately incorporated during training to improve the discriminative capability of the CNN. By explicitly exposing the network to visually confusing non-breakout structures, the CNN adapts to better distinguish true breakouts from similar non-breakout features, thereby reducing misidentification.

The use of dual-channel inputs allows the CNN to jointly exploit textural information from acoustic amplitude logs and geometric information from borehole radius logs. This complementary input representation is intended to improve robustness in breakout segmentation, particularly in cases where amplitude data alone may be insufficient to discriminate breakouts from non-breakout features, as shown in Figure 5. Within the Breakout-picker framework, the CNN aims at the identification of candidate breakout regions with high sensitivity, providing a reliable segmentation basis for subsequent determination of breakout parameters.

\subsubsection{Determination of breakout parameters}
After training, the CNN is applied to target boreholes, taking the acoustic amplitude log and borehole radius log as inputs. The CNN outputs a binary breakout segmentation image, which serves as the basis for determining candidate breakout parameters, including breakout azimuths and breakout widths.

\begin{figure*}[t]
\centering
\includegraphics[width=0.8\textwidth]{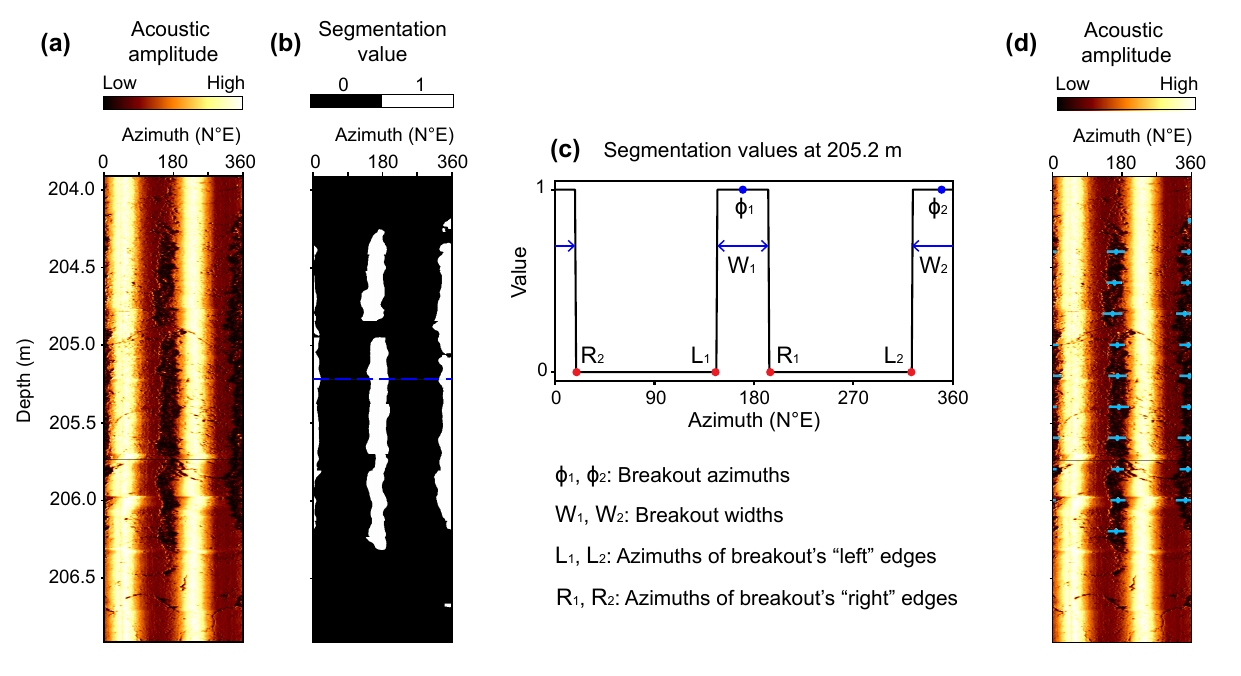}
\caption{Illustration of determining breakout parameters from the breakout segmentation image. (a) An acoustic amplitude log segment with borehole breakouts. (b) Binary breakout segmentation image. (c) Pixel values from the blue dashed line in (b), which are used to determine breakout azimuths, widths, and edges. (d) Breakout azimuths (cyan dots) and widths (cyan bars) on top of the acoustic amplitude log.}
\label{fig2}
\end{figure*}

The extraction of breakout parameters is performed depth by depth based on the spatial distribution of segmented breakout pixels. As shown in Figure \ref{fig2}, for a given measured depth, the breakout segmentation image (Figure \ref{fig2}b) can be represented as a one-dimensional binary function of azimuth (Figure \ref{fig2}c), where pixel values of one indicate breakout regions and pixel values of zero indicate non-breakout regions. At each depth, contiguous azimuthal intervals with pixel values equal to one are identified as individual breakout regions. For each breakout region $i$, the azimuth at which the pixel value transitions from zero to one is defined as the left edge, denoted as $L_i$, and the azimuth at which the pixel value transitions from one to zero is defined as the right edge, denoted as $R_i$. 

The breakout width $W_i$ is defined as the azimuthal extent of the breakout region, calculated as the clockwise angular distance from $L_i$ to $R_i$. A minimum width threshold of 10° is applied to exclude narrow segmented features that are unlikely to represent true borehole breakouts. Such narrow features are commonly associated with image noise, minor spalling, natural fractures, or segmentation artifacts, and their exclusion helps suppress spurious detections while retaining physical meaningful breakout regions. The breakout azimuth $\phi_i$ is defined as the midpoint between $L_i$ and $R_i$ in azimuth. This procedure is repeated for all measured depths to obtain depth-dependent candidate breakout azimuths and widths (Figure \ref{fig2}d), which are subsequently used for breakout validation and refinement.

\subsubsection{Breakout validation using azimuthal symmetry criteria}
Despite negative-sample-aware training, some non-breakout features may still be misidentified as breakouts by the CNN due to the limitation of training sample's diversity and network's capability. An example is shown in Figure ~\ref{fig1}, where a large-opening fracture is incorrectly segmented as a breakout. To further suppress such false positives, a rule-based breakout validation step is applied to the candidate breakout parameters extracted from the CNN's output. The validation criteria are adapted from the breakout determination criteria proposed by Wang and Schmitt \cite{wang2020automated} and are based on the characteristic azimuthal symmetry of borehole breakouts. Specifically, true breakouts typically occur as paired features on opposite sides of the borehole wall, reflecting the stress concentration around the borehole.

First, two candidate breakout azimuths are expected to occur at the same measured depth. Any depth interval with fewer or more than two candidate azimuths is considered inconsistent with typical breakout behavior and is therefore classified as a false positive and removed. Second, for depth intervals with two candidate breakout azimuths, denoted as $\phi_1$ and $\phi_2$, the azimuthal separation is required to satisfy:
\begin{equation}
160^{\circ} \leq \mathrm{circ}(\phi_2 - \phi_1) \leq 200^{\circ},
\label{eq:azimuth_separation}
\end{equation}
where $\mathrm{circ}(\phi_2 - \phi_1)$ denotes the azimuthal deviation between $\phi_1$ and $\phi_2$. This criterion ensures that the two candidate azimuths are approximately $180^{\circ}$ apart, consistent with the near-symmetrical geometry expected for true borehole breakouts.

We acknowledge that the proposed breakout validation criteria are idealized and may not be applicable to all breakout scenarios. For example, breakouts initiating near the depth bottom of a borehole may develop predominantly on one side of the borehole wall \cite{ito1998stress}. In cases where the magnitudes of the horizontal stresses are approximately equal, three breakouts may form around the borehole wall with angular separations of roughly $120^{\circ}$ \cite{haimson2007micromechanisms, duan2016evolution}. In addition, damaged weak zones at the intersection of fractures and the borehole may cause deviations of breakout azimuths from ideal symmetry \cite{sahara2014impact}. Nevertheless, these cases represent special geological or stress conditions and are not representative of typical borehole breakout behavior commonly encountered in most boreholes. For the purpose of automatic breakout characterization, the proposed validation criteria are effective in suppressing false positives and improving the reliability of breakout identification. We note, however, that under certain special conditions, the possibility of excluding true breakouts cannot be entirely ruled out.

\subsection{Network architecture}
\begin{figure*}[t]
\centering
\includegraphics[width=0.8\textwidth]{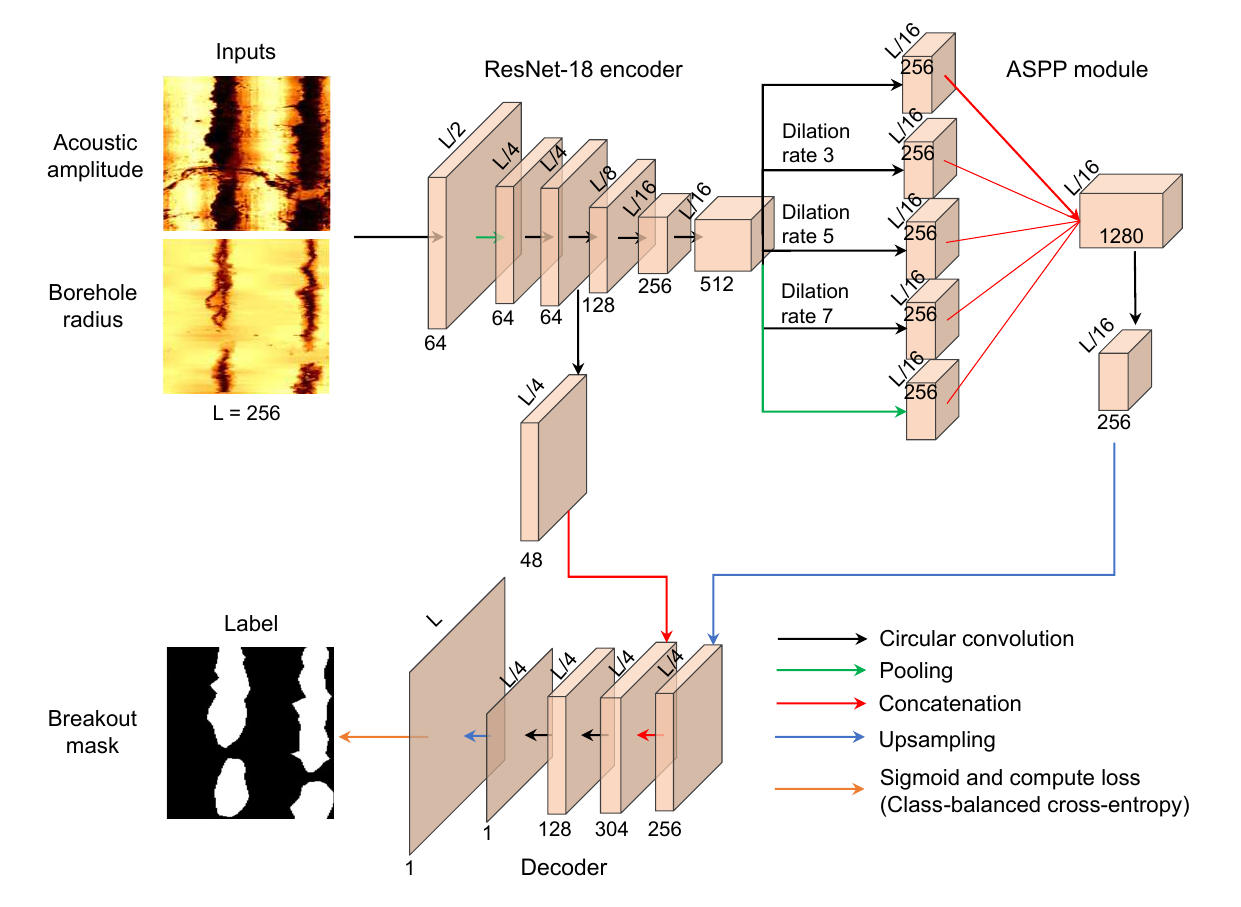}
\caption{Network architecture of Breakout-picker, which is adapted from DeepLabV3+ \cite{chen2018encoder}. The backbone is replaced with ResNet-18 instead of the original ResNet-101 to reduce model complexity and mitigate the risk of overfitting on small dataset. ASPP denotes the atrous spatial pyramid pooling modules. Circular convolution treats the input as periodic, so that the two azimuthal edges (0° and 360°) are connected rather than padded with zeros. The numbers annotated on the feature maps indicate the number of channels at each stage of the network. The labels L/2, L/4, L/8, and L/16 denote the relative square lengths of feature maps with respect to the input image’s square length L.}
\label{fig3}
\end{figure*}

Breakout-picker adopts an encoder-decoder architecture for segmenting breakouts from acoustic image logs. As shown in Figure \ref{fig3}, the network architecture is adapted from the DeepLabV3+ framework \cite{chen2018encoder}, with the original ResNet-101 backbone replaced by a lighter ResNet-18 to reduce model complexity and mitigate overfitting on the relatively small training set. The input consists of two channels: acoustic amplitude and borehole radius. These two channels provide complementary information, with amplitude capturing borehole wall reflectivity variations and radius highlighting geometric enlargement associated with breakouts. To respect the circular geometry of borehole images, circular convolution is applied along the azimuthal direction by enforcing a periodic boundary condition across the 0°/360° boundary.

The encoder follows the DeepLabV3+ feature hierarchy and progressively extracts representations from local textures to higher-level structural patterns through successive down-sampling stages. This hierarchical design is useful for borehole images because breakout signatures exhibit both fine-scale appearance (e.g., low acoustic amplitude zones with irregular textures) and larger-scale geometric characteristics (e.g., continuous damaged zones extending with depth). In addition, acoustic image logs often contain acquisition-related artifacts. Therefore, relying on a single receptive field may be insufficient to separate true breakouts from visually similar non-breakout features. To address this, the deepest encoder features are passed to an atrous spatial pyramid pooling (ASPP) module, which applies parallel atrous convolutions with multiple dilation rates to aggregate context at different effective receptive fields. By integrating information over multiple spatial scales, ASPP is expected to improve sensitivity to both narrow and wide breakouts and reduces sensitivity to variations in breakout size, boundary sharpness, and local image quality. The outputs of the parallel atrous branches are concatenated to form a multi-scale feature representation that combines fine-scale texture characteristics with larger-scale geometric patterns of breakout regions.

The decoder reconstructs dense predictions by progressively up-sampling the ASPP features and fusing them with low-level encoder features via skip connections. This fusion is important because down-sampling in the encoder improves semantic abstraction but inevitably loses boundary detail, while breakout parameter extraction later depends on accurate localization of breakout edges. The skip connections reintroduce higher resolution spatial information from earlier layers, allowing the decoder to refine the segmentation boundaries and better preserve the azimuthal extent of breakout regions. The decoder design thus balances contextual interpretation from deep features and precise delineation from shallow features, which is particularly relevant for acoustic images where breakout boundaries may be partially obscured by logging artifacts or intersecting textures such as fractures. The network terminates with a single-channel output followed by a sigmoid activation, producing a pixel-wise breakout probability map with values from 0 to 1.

The network is trained using a pixel-wise loss function applied to the predicted probability map. To account for the strong class imbalance between breakout and background pixels, we adopt the following class-balanced binary cross-entropy loss:
\begin{equation}
L = -\beta \sum_{i=0}^{N} y_i \log(p_i) - \sum_{i=0}^{N} (1 - y_i) \log(1 - p_i),
\end{equation}
where $N$ denotes the number of pixels in the input acoustic image log, $y_i$ represents the binary pixel values of the breakout labels, $p_i$ denotes breakout probabilities, and $\beta = \sum_{i=0}^{N}(1 - y_i) / N$ represents the ratio of the number of non-breakout pixels to the total number of pixels. In acoustic image logs, breakout regions typically occupy only a minor fraction of the borehole wall, and the use of a standard binary cross-entropy loss can potentially bias the network toward background prediction. The class-balanced loss assigns higher weight to breakout pixels, facilitating accurate identification of breakout regions while maintaining stable training.

Overall, the architecture of Breakout-picker extends the DeepLabV3+ framework to acoustic image log analysis by incorporating circular convolution and dual-channel inputs. Together with multi-scale feature aggregation and class-balanced loss, these design choices are intended to enable reliable and automatic breakout segmentation from acoustic image logs.

\section{Dataset}
\subsection{Overview of breakout samples}
In this study, a breakout dataset was constructed from acoustic image logs acquired from the Bedretto Underground Laboratory (BedrettoLab) in Switzerland. BedrettoLab is an underground geoscience research facility established by ETH Zürich in 2018 within the Bedretto Tunnel in Ticino, Switzerland, designed as a hectometer-scale in situ test-bed for multidisciplinary experiments in fractured crystalline rock under ~1000 m overburden \cite{ma2022multi}. An array of eight sub-parallel boreholes, with lengths ranging from 192 m to 405 m, were drilled into a granitic rock mass for in-situ experiments and are fully covered by acoustic image logs.

The acoustic image logs from these eight boreholes reveal abundant borehole breakouts, which have previously been used to analyze fault zone-related stress variations \cite{zhang2023fault}. However, the image quality and logging responses are not uniform across the boreholes. Variations arise from differences in drilling methods, borehole wall conditions, and logging environments, which influence acoustic scattering behavior, noise characteristics, and the appearance of logging artifacts. 

\begin{figure*}[t]
\centering
\includegraphics[width=0.8\textwidth]{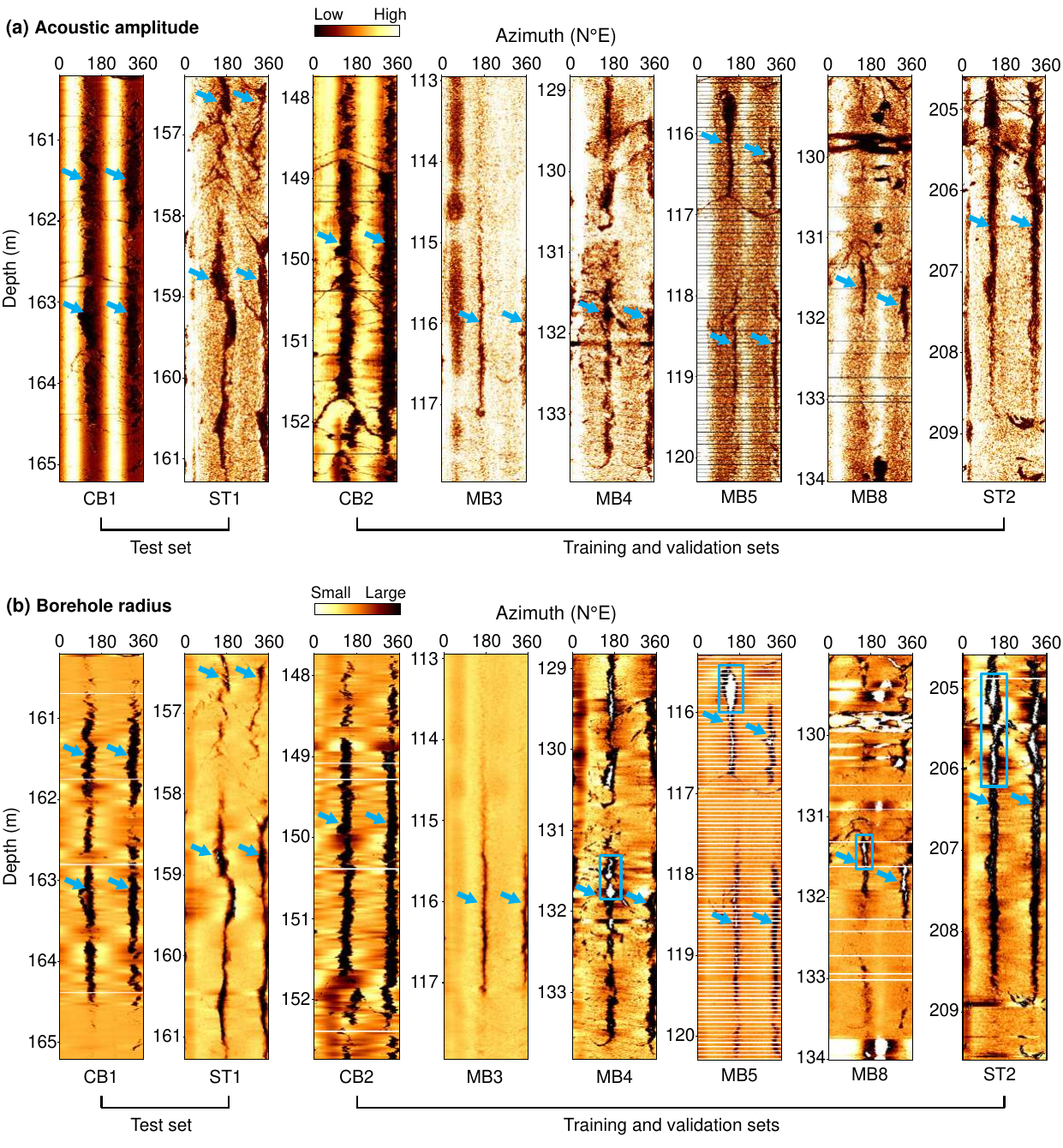}
\caption{Examples of borehole breakouts (indicated by the blue arrows) in (a) the acoustic amplitude logs and (b) borehole radius logs from eight boreholes at the BedrettoLab. Acoustic image logs from borehole CB2, MB3, MB4, MB5, MB8, and ST2 are used to construct training and validation sets, while acoustic image logs from borehole CB1 and ST1 are used as the test set. Note that boreholes CB1 and CB2 were later used for monitoring purpose, which are also referred to as CB1 and CB2. The blue boxes in the borehole radius logs of MB4, MB5, MB8, and ST2 indicate anomalously small radii within breakout regions.}
\label{fig4}
\end{figure*}

For example, as shown in Figure \ref{fig4}a, the acoustic amplitude logs of CB1 and CB2 exhibit noticeably less speckle noise compared to the other boreholes, likely manifesting smoother borehole wall conditions. These two boreholes were drilled using coring bits, whereas the remaining boreholes were drilled using hammer bits. Hammer-bit drilling may leave a more irregular borehole wall than coring, increasing ultrasonic scattering and thereby degrading acoustic image quality \cite{blake2012borehole, thomas2015structural}. Consequently, the dataset reflects realistic field conditions rather than uniformly high-quality images, providing a challenging and representative testbed for automatic breakout characterization.

Representative breakouts examples from the eight boreholes are illustrated in Figure \ref{fig4}. Breakouts observed in boreholes CB1 and CB2 are wider than those in other boreholes, which may be attributed to lower strength of the surrounding rock mass or higher stress concentration on the borehole wall \cite{zoback1985well}. In addition to breakouts, several logging-related artifacts are present in the dataset. A pair of prominent dark vertical stripes appears in the acoustic amplitude logs of CB1 and CB2, which are likely caused by tool eccentricity during logging. Moreover, although breakouts are typically associated with enlarged borehole radii, anomalously small radii are observed within some breakout regions, such as in boreholes MB4, MB5, MB8, and ST2. These features are interpreted as another type of logging artifact, potentially resulting from automatic signal compensation in regions with severe acoustic signal loss caused by highly irregular breakout surfaces. For model evaluation, the complete acoustic image logs from boreholes CB1 and ST1 are designated as the test set, while the remaining boreholes are used to construct the training and validation datasets.

\subsection{Breakout annotation}
\begin{figure*}[t]
\centering
\includegraphics[width=0.8\textwidth]{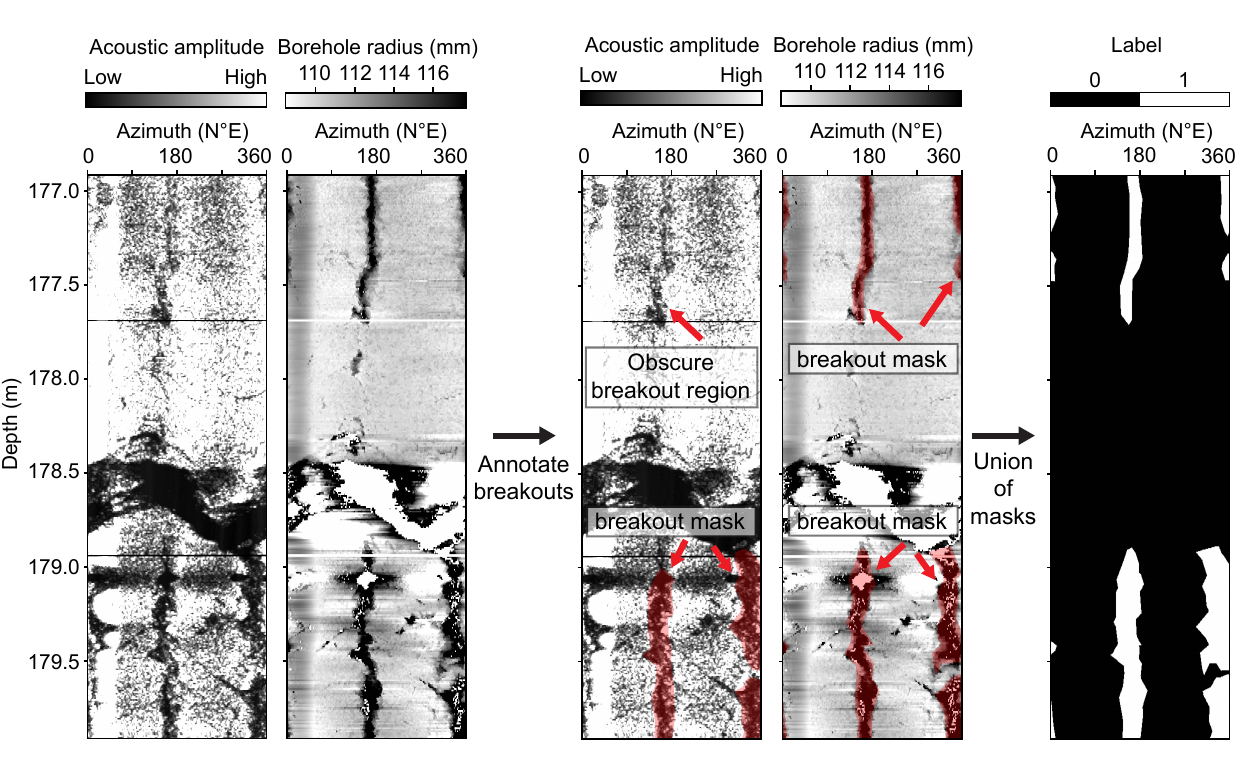}
\caption{Breakout annotation workflow based on acoustic amplitude and borehole radius logs. The example is taken from borehole ST2. Breakout masks (red) are independently annotated on the acoustic amplitude and borehole radius logs, and the final breakout mask is obtained by taking the union of the two annotations.}
\label{fig5}
\end{figure*}

Breakout annotation is performed using both acoustic amplitude and borehole radius logs as complementary data sources (Figure \ref{fig5}). The two measurements respond to different physical aspects of breakout development: acoustic amplitude primarily reflects changes in borehole wall roughness and reflectivity, whereas borehole radius measurements capture geometric enlargement of the borehole cross-section. Because breakouts develop as zones of wall damage rather than ideal sharp cavities, their expression can differ between the two logs. The annotation procedure therefore does not assume one modality to be intrinsically more accurate than the other.

In practice, the apparent breakout extent may differ between amplitude and radius logs. Acoustic amplitude can show a broader low-amplitude zone where the borehole wall is roughened, fractured, or mechanically weakened, even if part of that material remains in place. Radius logs, on the other hand, emphasize geometric enlargement and may highlight zones where material has been removed but where amplitude contrast is reduced by noise, signal attenuation, or logging artifacts. These differences reflect measurement sensitivity rather than contradiction.

To account for these complementary sensitivities, breakout masks are annotated independently on amplitude and radius logs, and the final breakout label is defined as the union of the two masks (Figure \ref{fig5}). This strategy is intended to reduce omission of breakouts that are clearly expressed in one modality but partially obscured in the other, for example by speckle noise in amplitude images or weak geometric contrast in radius data. The union therefore represents an integration of roughness-related and geometry-related evidence, rather than a preference for either measurement.

Intervals where the two logs suggest markedly different breakout extents are not interpreted as conflicts, but as cases where breakout signatures are more clearly resolved in one measurement than in the other. Because the goal of annotation is to provide inclusive training labels for segmentation, retaining breakout regions visible in either log is preferable to discarding potentially valid breakout zones. This dual-log integration approach is consistent with practical breakout picking workflows, where image-based indicators of wall damage and caliper or radius-based indicators of borehole enlargement are considered jointly when delineating breakout zones.

\subsection{Negative samples}
\begin{figure*}[t]
\centering
\includegraphics[width=0.8\textwidth]{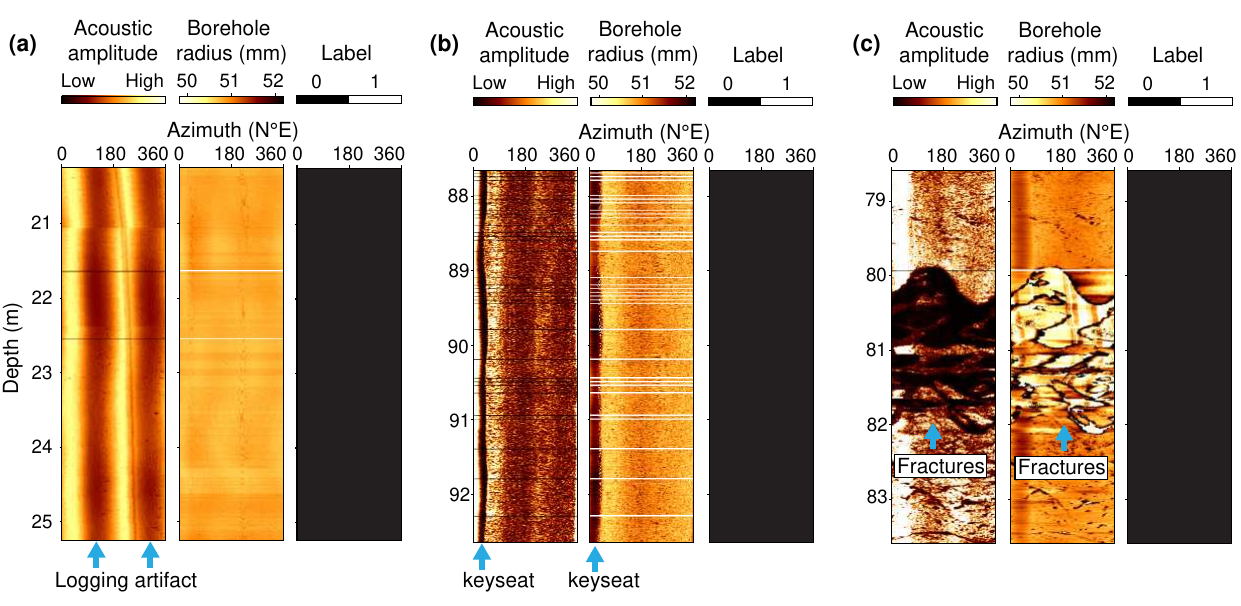}
\caption{Non-breakout areas in acoustic image logs that share similar features with breakouts. (a) Low-amplitude logging artifacts in CB2, (b) keyseats in ST2, and (c) large-opening (aperture $\ge$ 10 cm) fractures in MB4. Segmentation labels for these non-breakout areas are all zeros, indicating the absence of breakouts.}
\label{fig6}
\end{figure*}

In addition to borehole breakouts, we observed several non-breakout features in the acoustic image logs. These features resemble breakout signatures and can potentially be a major source of false positives in automatic breakout characterization. These features include logging artifacts, keyseats, and large-aperture fractures, all of which may present low acoustic amplitude, apparent borehole wall damage, or azimuthal patterns resembling those of breakouts. Representative examples of these non-breakout features are shown in Figure \ref{fig6}.

Figure \ref{fig6}a illustrates a logging artifact observed in the ATV logs of borehole CB2, characterized by a pair of vertical low-amplitude stripes that are approximately 180° apart in azimuth. Although this pattern closely resembles the paired geometry of borehole breakouts, it is not accompanied by a corresponding increase in borehole radius. Figure \ref{fig6}b shows a keyseat in borehole ST2. A keyseat is an unintended groove that develops on the borehole wall during drilling, caused by abnormal drill pipe motion such as lateral vibration or eccentric rotation, which leads to repeated contact and friction between the rotating drill string and the borehole wall. Keyseats share similar acoustic amplitude reduction and geometric appearance with breakouts, but they typically occur on only one side of the borehole, in contrast to the paired development of breakouts. Figure \ref{fig6}c presents large-opening (aperture $\ge$ 10 cm) fractures intersecting borehole MB4. These fractures appear as sinusoidal low-amplitude bands in the acoustic images and are frequently observed across the studied boreholes. Although such fractures may theoretically correspond to local borehole enlargement, they instead appear as regions of reduced borehole radius in the radius logs, likely due to logging-related effects such as signal loss and automatic signal compensation over highly irregular fracture surfaces, which could be resulted from internal material fall-off after drilling.

These non-breakout features are deliberately included as negative samples during Breakout-picker training. Rather than using randomly selected background regions, which are typically easy to classify and contribute little to improving model discrimination, we focus on these hard negative samples that closely resemble true breakouts in appearance. By explicitly exposing the Breakout-picker to the primary sources of false positives, the network is encouraged to learn more discriminative feature representations that better separate true breakouts from visually similar non-breakout structures.

\subsection{Sample augmentation}
\begin{figure*}[t]
\centering
\includegraphics[width=0.8\textwidth]{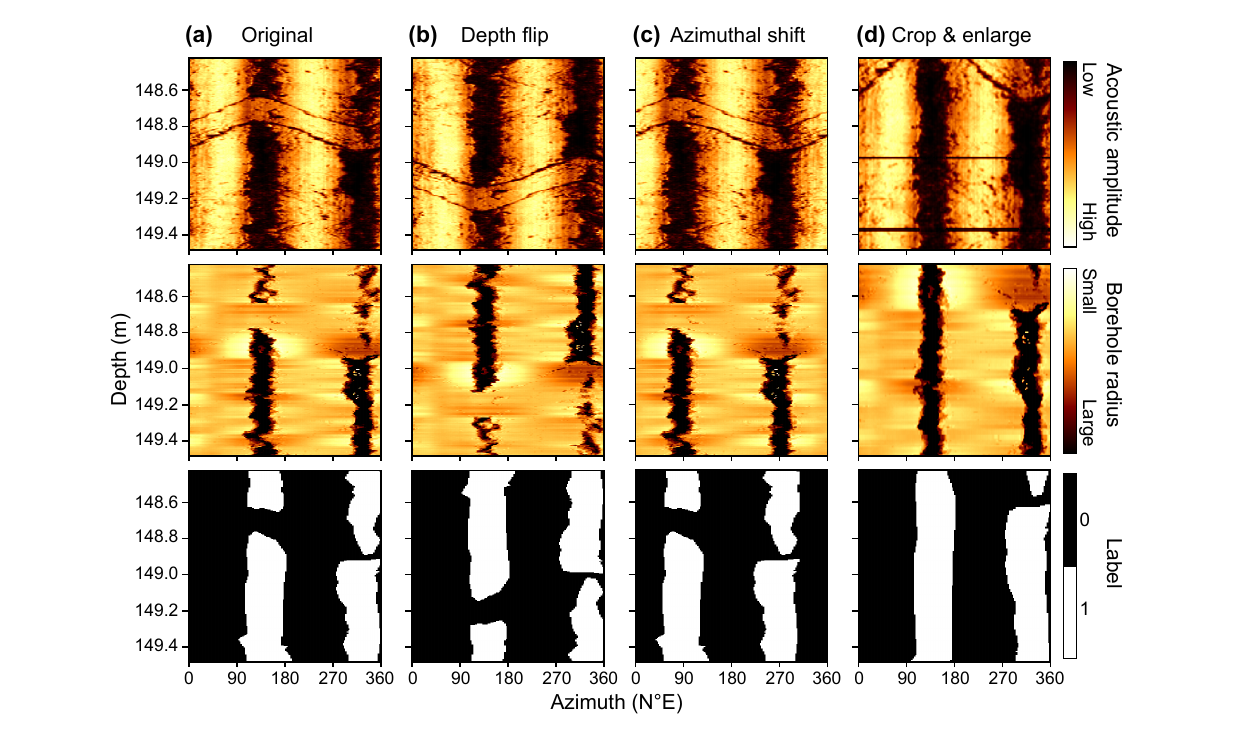}
\caption{Illustration of sample augmentation used to increase the size and diversity of the training dataset. (a) An original sample from borehole CB2, including the acoustic amplitude log, borehole radius log, and corresponding breakout label, with a size of 256 × 256 pixels. The sample is augmented by (b) depth flipping, (c) azimuthal shifting, and (d) random cropping followed by enlarging.}
\label{fig7}
\end{figure*}

To enhance the diversity of training samples, a set of data augmentation strategies is applied to both positive and negative samples. Each sample has a size of 256 × 256 pixels in the depth-azimuth coordinate, consisting of the dual-channel inputs (acoustic amplitude and borehole radius logs) together with the corresponding breakout label. Figure \ref{fig7} illustrates the augmentation operations, including depth flipping, azimuthal shift, and crop-and-enlarge. For each augmented sample, the same geometric transformation is applied consistently to both input channels and the label mask, ensuring that the spatial correspondence between amplitude, radius, and breakout annotation is preserved.

Depth flipping reverses the depth axis of the image patch. This operation increases variability in the vertical arrangement of breakout intervals and their surrounding context, while preserving breakout geometry in azimuth. It intends to reduce the risk that the network implicitly associates breakout patterns with a fixed depth-wise ordering of textures or neighboring structures within the patch, thereby improving robustness to depth-dependent variations in breakout appearance.

Azimuthal shifting moves the image sideways along the azimuthal direction, with wrap-around at the 0°/360° boundary. It can redistribute breakouts to different absolute azimuthal positions. This operation increases the diversity of breakout samples in terms of where breakouts are located along the borehole circumference. Without this augmentation, the network may implicitly learn that breakouts tend to occur at certain pixel columns simply because of the original azimuthal alignment in the dataset. Azimuthal shifting breaks this artificial positional bias and encourages the model to focus on the intrinsic visual and geometric characteristics of breakouts, rather than their fixed positions in the training images. This helps the trained model generalize better to new boreholes where breakouts may appear at different azimuths due to changes in in-situ stress orientation or borehole trajectory. In practice, azimuthal shifting is applied twice to each sample, with a random shift angle uniformly drawn from 45° to 135° each time.

In addition to these global transformations, we apply a random crop-and-enlarge operation to positive samples. A subregion that contains breakout pixels (according to the label mask) is randomly cropped and then resized back to 256 × 256 pixels. This augmentation increases variability in breakout scale and relative proportion between breakout and background regions, which encourages the model to recognize breakouts across a broader range of apparent widths and geometric extents rather than relying on fixed-scale patterns. The crop-and-enlarge operation is not applied to negative samples, because cropping may inadvertently remove the non-breakout structures of interest or distort their characteristic geometry, weakening their role as hard negatives.

All augmentation operations except crop-and-enlarge are applied to both positive and negative samples. Through these strategies, the number of positive samples is increased from 135 to 675, and the number of negative samples from 93 to 465. Collectively, these augmentations expand diversity along depth context, azimuthal position, and breakout scale, improving the generalization of Breakout-picker across varying breakout appearances.

\section{Results}
\subsection{Breakout segmentation}
To evaluate the performance of the proposed Breakout-picker in borehole breakout segmentation, we compare its results with those obtained using MMDC-UNet, a DL-based breakout characterization approach proposed by Yang et al. \cite{yang2022automated}. MMDC-UNet is a semantic segmentation network built upon the U-Net architecture and enhanced with densely connected multi-module blocks. In the study of Yang et al. \cite{yang2022automated}, the MMDC-UNet was trained to segment breakouts from acoustic amplitude logs using only positive breakout samples. It achieved an average accuracy of 93.79\% (i.e., the proportion of correct pixel labels in predicted images) over two unseen datasets, demonstrating the ability to accurately delineate breakout regions from acoustic amplitude logs.

To ensure a fair comparison and to isolate the effect of training strategy, MMDC-UNet is implemented in this study using the same dual-channel inputs as Breakout-picker, namely acoustic amplitude and borehole radius logs. Apart from the input configuration, the original training paradigm of MMDC-UNet is retained, in which the network is trained exclusively on positive samples corresponding to manually annotated breakout regions. In contrast, Breakout-picker adopts an alternative strategy by incorporating both positive and hard negative samples during training.

\begin{figure*}[t]
\centering
\includegraphics[width=0.8\textwidth]{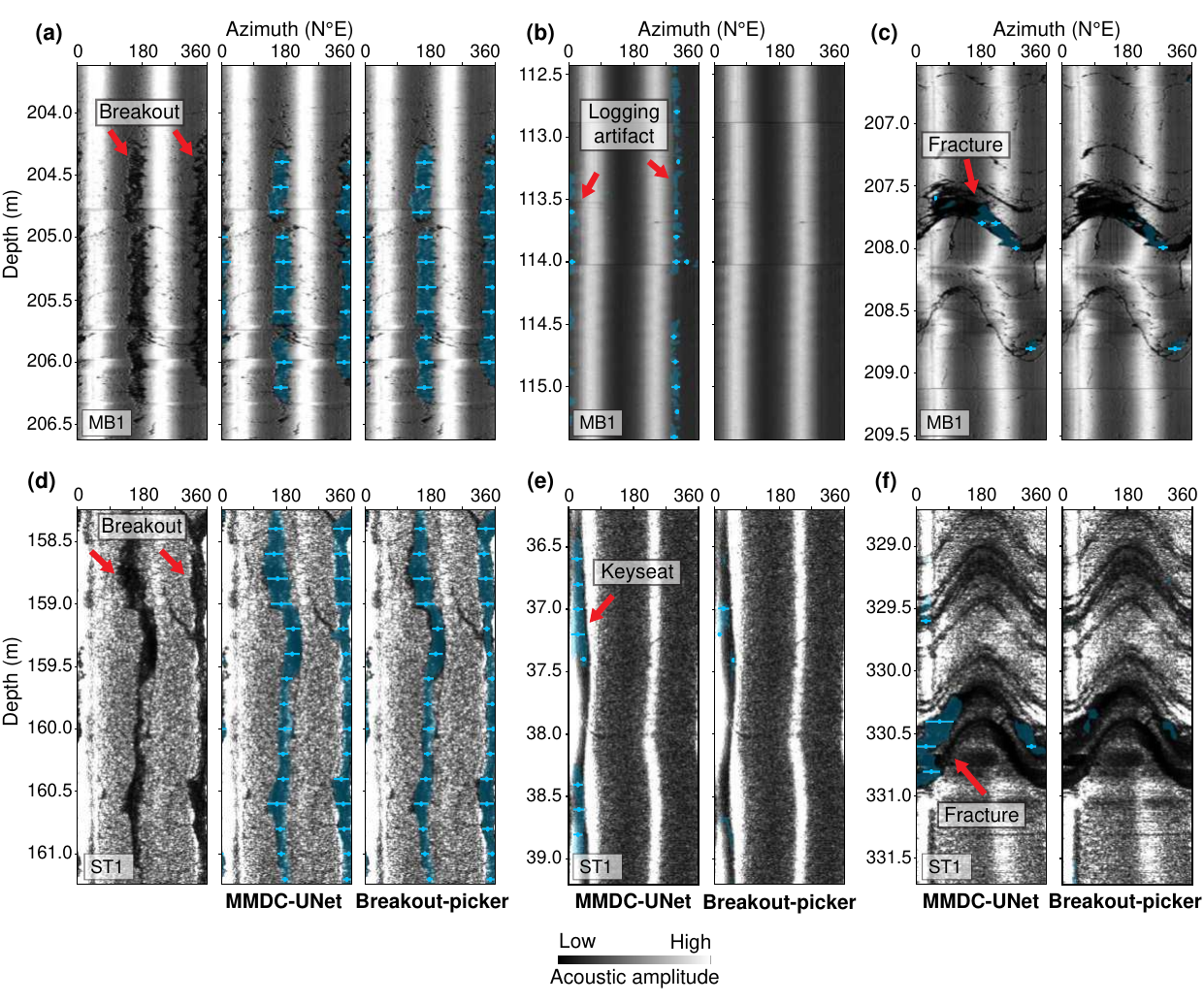}
\caption{Qualitative comparison of breakout segmentation results between MMDC-UNet and Breakout-picker. (a – c) Results from CB1 and (d – f) from ST1. Blue regions denote predicted breakout segmentation, overlaid by breakout azimuths (cyan dots) and widths (cyan bars) determined from the breakout segmentation results.}
\label{fig8}
\end{figure*}

A qualitative comparison of breakout segmentation results produced by MMDC-UNet and Breakout-picker in Figure \ref{fig8} for several representative cases, including true breakouts, logging artifacts, keyseats, and fracture-dominated intervals. For true borehole breakouts, both methods are able to accurately identify the main breakout regions along the borehole wall, producing segmentation results that are generally consistent with the actual breakout regions. The segmentation outputs from both methods capture the overall geometry and spatial extent of the breakouts, indicating that DL-based approaches are effective for identifying true breakouts in acoustic image logs.

Pronounced differences between the two methods are observed for non-breakout features. Logging artifacts, which often appear as vertical low-amplitude stripes in acoustic amplitude logs, are frequently misidentified as breakouts by MMDC-UNet. Similarly, large-opening fractures and keyseats are commonly segmented as breakout regions due to their low acoustic amplitude or locally enlarged borehole geometry. These misclassifications reflect the inherent limitation of segmentation models trained solely on positive samples, where visually similar but physically distinct features are not explicitly penalized during training.

In comparison, Breakout-picker effectively suppresses these false positives across different non-breakout scenarios. Logging artifacts, fractures, and keyseats are largely excluded from the predicted breakout regions, while true breakouts remain consistently detected. This improved discrimination capability demonstrates the benefit of incorporating hard negative samples into the training process, enabling the network to learn more robust decision boundaries between breakout and non-breakout features.

In addition to the qualitative comparison, we further evaluate the segmentation performance of the two methods using the intersection-over-union (IoU) metric computed over the entire borehole length for the two test boreholes CB1 and ST1. The IoU is calculated between the network-predicted breakout segmentation and the manually annotated breakout labels for each borehole. As summarized in Table \ref{tab1}, for borehole CB1, MMDC-UNet achieves an IoU of 0.54, whereas Breakout-picker yields a substantially higher IoU of 0.72. For borehole ST1, the IoU values are 0.40 for MMDC-UNet and 0.45 for Breakout-picker, respectively. These results indicate that, although both methods are capable of segmenting true breakout regions, Breakout-picker consistently achieves higher agreement with manual annotations due to the suppression of false-positive breakout detections.

\begin{table*}[!t]
\centering
\caption{Quantitative comparison of breakout characterization performance for different automatic methods.}
\label{tab1}
\begin{tabular}{llccccccc}
\toprule
\textbf{Borehole} & \textbf{Method} & \textbf{IoU} & \textbf{Breakout azimuth ($^\circ$)} & \textbf{Azimuth error ($^\circ$)} & \textbf{Width error ($^\circ$)} & \textbf{FPR} & \textbf{FNR} \\
\midrule
\multirow{4}{*}{\shortstack[l]{CB1\\(143$^\circ$ $\pm$ 17$^\circ$)}}
& Peak Detection & -- & 147 $\pm$ 25 & 5.71 & 14.56 & 39\% & 14\% \\
& MMDC-UNet & 0.54 & 144 $\pm$ 28 & 3.90 & 14.54 & 48\% & 10\% \\
& Breakout-picker (Before validation) & 0.72* & 143 $\pm$ 18 & 3.46 & 10.32 & 14\% & 2\%* \\
& Breakout-picker (After validation) & 0.72* & 147 $\pm$ 14 & 2.29* & 7.03* & 4\%* & 48\% \\
\midrule
\multirow{4}{*}{\shortstack[l]{ST1\\(163$^\circ$ $\pm$ 17$^\circ$)}}
& Peak Detection & -- & 175 $\pm$ 31 & 5.17 & 32.54 & 67\% & 39\% \\
& MMDC-UNet & 0.40 & 175 $\pm$ 33 & 4.21 & 7.35* & 54\% & 5\%* \\
& Breakout-picker (Before validation) & 0.45* & 170 $\pm$ 31 & 3.13 & 10.43 & 39\% & 11\% \\
& Breakout-picker (After validation) & 0.45* & 166 $\pm$ 12 & 1.95* & 8.57 & 3\%* & 51\% \\
\midrule
\multirow{4}{*}{\shortstack[l]{Hunt Well\\(125$^\circ$ $\pm$ 43$^\circ$)}}
& Peak Detection & -- & 117 $\pm$ 39 & 7.27 & 21.07 & 37\% & 62\% \\
& MMDC-UNet & 0.42 & 119 $\pm$ 45 & 6.67 & 15.90 & 46\% & 20\% \\
& Breakout-picker (Before validation) & 0.49* & 119 $\pm$ 44 & 6.56 & 13.43 & 47\% & 10\%* \\
& Breakout-picker (After validation) & 0.49* & 123 $\pm$ 42 & 5.63* & 12.09* & 28\%* & 46\% \\
\midrule
\multirow{4}{*}{\shortstack[l]{IODP 1256D\\(93$^\circ$ $\pm$ 42$^\circ$)}}
& Peak Detection & -- & 91 $\pm$ 47 & 8.45 & 18.24 & 39\% & 68\% \\
& MMDC-UNet & 0.41 & 84 $\pm$ 47 & 8.84 & 19.60 & 49\% & 6\%* \\
& Breakout-picker (Before validation) & 0.44* & 87 $\pm$ 43 & 8.84 & 18.96 & 47\% & 10\% \\
& Breakout-picker (After validation) & 0.44* & 100 $\pm$ 36 & 6.59* & 14.90* & 14\%* & 71\% \\
\bottomrule
\end{tabular}

\vspace{4pt}
\raggedright
\footnotesize
\textit{Note:} Manually picked breakout azimuths (mean $\pm$ standard deviation) are denoted below borehole names. Azimuth and width errors denote the mean breakout azimuth and width errors between the results of manual picking and automatic methods. IoU denotes the intersection-over-union score between manual and automatic breakout segmentation. FPR and FNR denote the false positive rate and false negative rate, respectively. All breakout azimuths are relative to magnetic north. Best-performing metrics are marked by asterisks (*).
\end{table*}

Overall, the qualitative results indicate that Breakout-picker achieves more reliable breakout segmentation than MMDC-UNet, particularly in reducing false positives associated with visually ambiguous non-breakout structures. Such improvements in segmentation reliability provide a more stable basis for subsequent extraction of breakout parameters.

\subsection{Characterization of breakout parameters}
This section evaluates the capability of the proposed Breakout-picker framework to automatically characterize key breakout parameters, with a particular focus on breakout azimuth and breakout width. The performance of Breakout-picker is compared against two commonly used automatic approaches: the Peak Detection method based directly on acoustic image responses \cite{wang2020automated}, and the MMDC-UNet framework \cite{yang2022automated}. All comparisons are conducted on the two test boreholes, CB1 and ST1, using a uniform sampling interval of 0.2 m along the borehole depth.

\subsubsection{Breakout azimuth extraction}
\begin{figure*}[t]
\centering
\includegraphics[width=0.6\textwidth]{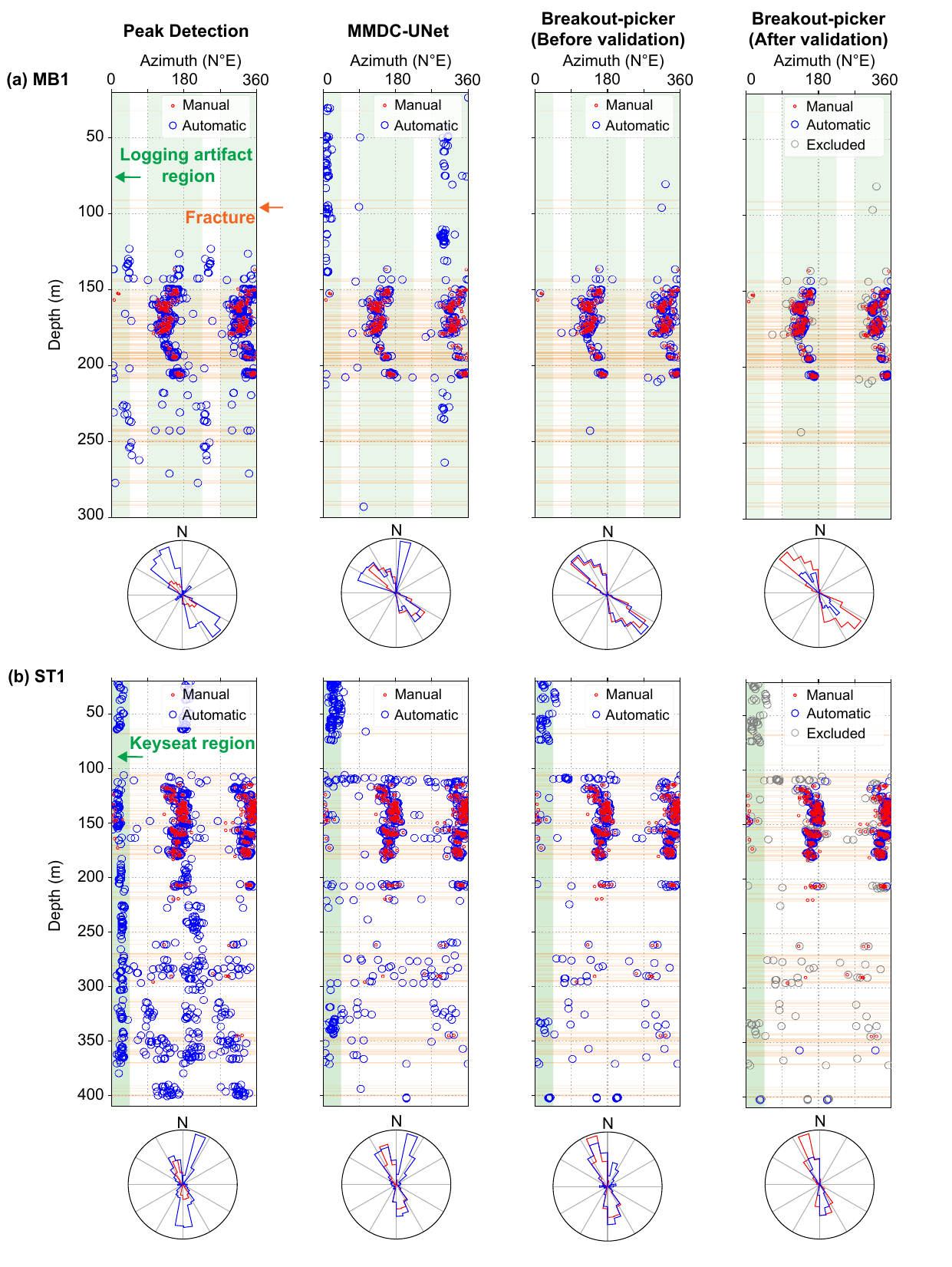}
\caption{Comparison of breakout azimuths determined using different methods for (a) CB1 and (b) ST1. Red and blue circles denote manually and automatically picked breakout azimuths. Gray circles represent azimuths excluded by Breakout-picker’s validation criteria. The green vertical bands indicate the low-amplitude logging artifact regions in CB1 and keyseat regions in ST1. The yellow transverse lines indicate the fracture locations. The rose diagrams plot the azimuthal distributions of manually (red) and automatically (blue) picked breakouts.}
\label{fig9}
\end{figure*}

Breakout azimuths extracted from CB1 and ST1 using different automatic methods are compared in Figure \ref{fig9}. For the Peak Detection approach, breakout azimuths are directly inferred from characteristic responses in acoustic amplitude and borehole radius logs, whereas for MMDC-UNet and Breakout-picker, azimuths are derived from the segmented breakout regions. 

For CB1, the manually interpreted breakouts show a predominant NW-SE orientation. As shown in the rose diagrams in Figure \ref{fig9}a, the breakout azimuths determined by Breakout-picker (before validation) accurately capture both the primary direction and the directional distribution of the true breakouts. After applying the validation criteria, the directional distribution becomes less representative of the true breakout population, but the primary NW-SE orientation is still correctly identified. For ST1, the true breakouts are predominantly in the NNW-SSE direction. Breakout-picker also correctly identifies this primary orientation. In contrast to CB1, the validation process here improves the results. By excluding false detections in keyseat regions, the directional distribution better represents the true breakout population.

Overall, all three automatic approaches are able to recover the main breakout trends within depth intervals where well-developed paired breakouts are present. The breakout azimuth errors between manual picking and the three automatic approaches are all less than 6° for both CB1 and ST1 (Table \ref{tab1}). However, marked differences arise in the amount and spatial distribution of false detections. The Peak Detection method produces a large number of spurious azimuth picks, especially in ST1, where numerous non-breakout features such as keyseats and fractures are present. These false positives introduce a spurious peak in the rose diagram, potentially misleading stress direction interpretation. The false positive rate (FPR) of Peak Detection method is 39\% for CB1 and reaches an even higher value of 67\% for ST1. These spurious picks result from the direct reliance on local amplitude or radius anomalies without an explicit breakout segmentation step. Keyseats and fractures could also exhibit local amplitude or radius anomalies that satisfy the breakout determination criteria of Peak Detection method, leading to misidentifications and elevated FPR.

The MMDC-UNet approach reduces the azimuth error by a small margin (1° for CB1 and 1.8° for ST1) compared to Peak Detection, which could be due to the breakout azimuths are better constrained by the segmented breakout regions. Nevertheless, MMDC-UNet still produces a considerable number of isolated azimuthal picks outside manually identified breakout intervals. The FPR for CB1 is even increased by 9\%. This behavior reflects the fact that MMDC-UNet represents a class of DL-based breakout characterization methods that are trained exclusively on positive breakout samples. As a result, non-breakout structures that share visual similarities with breakouts, such as keyseats, large-opening fractures, or logging artifacts, may still be misclassified as breakouts and subsequently contribute spurious azimuth estimates, as demonstrated in Figure \ref{fig8}.

By contrast, the Breakout-picker framework yields a noticeably cleaner azimuth distribution in both CB1 and ST1, especially after applying the breakout validation criteria. Prior to validation, the Breakout-picker azimuths already show improved agreement with manual picks compared to MMDC-UNet. As shown in Table \ref{tab1}, compared to the MMDC-UNet approach, the FPR of Breakout-picker on CB1 reduced from 48\% to 14\%, and from 54\% to 39\% on ST1. This reduction in FPR reflects the benefit of explicitly incorporating negative samples during training to suppress false breakout segmentation. The breakout azimuth error remains as low as about 3° for both CB1 and ST1, indicating that including negative training samples does not degrade the segmentation accuracy at breakout regions. 

After validation, many remaining isolated azimuth picks are further removed, resulting in a distribution that more closely follows the manually interpreted breakout pairs. Quantitatively, the FPR after validation reduced from 14\% to 4\% for CB1 and from 39\% to 3\% for ST1, indicating a more reliable breakout azimuth characterization with almost no false detection. However, it should be noted that this reduction in false positives is accompanied by an increase in the false negative rate (FNR) after applying the breakout validation criteria. The FNR increases by more than 40\% for both CB1 and ST1. This observation indicates that a number of true breakouts are excluded by the breakout validation criteria. The underlying causes of this trade-off are examined in detail in the following section.

\subsubsection{Effect of breakout validation criteria}
\begin{figure*}[t]
\centering
\includegraphics[width=0.8\textwidth]{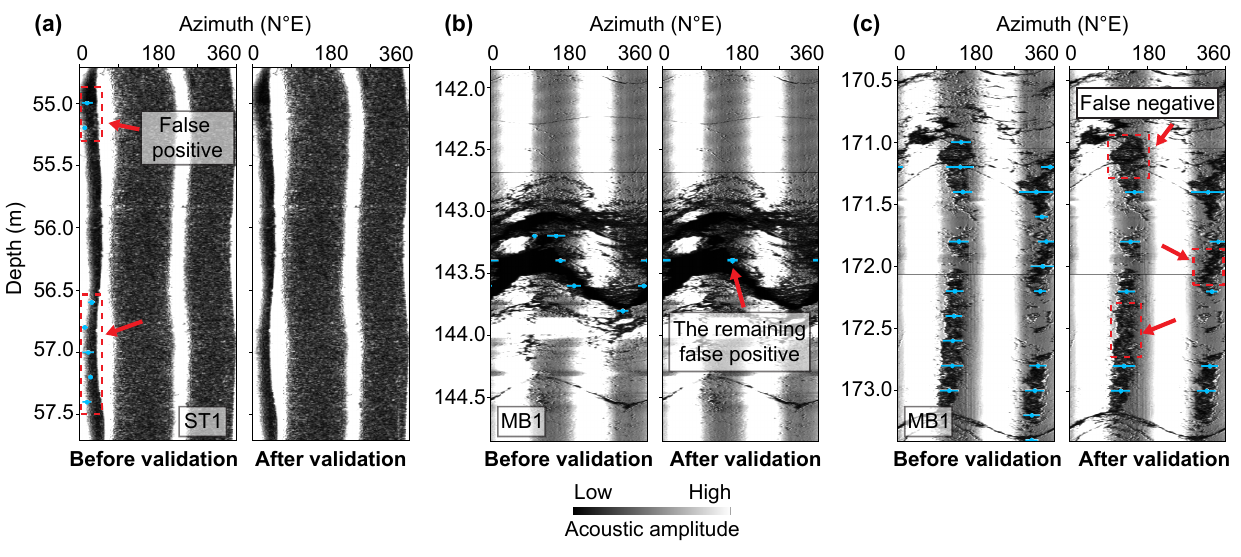}
\caption{Examples showing the effect of applying the breakout validation criteria. (a) False breakout detections over a keyseat are completely removed. (b) False breakout detections over a large-opening fracture are reduced but several false positives persist. (c) True breakout detections are partially excluded due to the absence of a paired breakout on the opposite borehole wall, resulting in false negatives. }
\label{fig10}
\end{figure*}

A detailed comparison of breakout azimuth picks before and after applying the breakout validation criteria is provided in Figure \ref{fig10}, focusing on three representative depth intervals. These examples illustrate the dual role of the validation step in suppressing false positives while potentially introducing additional false negatives.

The first example (Figure \ref{fig10}a) shows a keyseat interval in borehole ST1 at measured depths of 55 to 57.5 m. Before validation, the keyseat is incorrectly interpreted as breakouts due to its elongated geometry and low acoustic amplitude. Because these azimuth picks do not form a near-diametrically opposite pair, they are effectively removed after applying the validation criteria. This example demonstrates the ability of the validation step to eliminate non-breakout features that are geometrically inconsistent with stress-induced breakouts.

A fracture-dominated interval in borehole CB1 at measured depths of 143 to 144 m is highlighted in Figure \ref{fig10}b, where a large-opening fracture produces several false breakout detections. After validation, most of these spurious picks are removed. However, a small number of false positives remain because their azimuths accidentally satisfy the near-symmetric criterion. This case illustrates an inherent limitation of symmetry-based validation. While effective in most situations, it cannot completely eliminate all false positives when non-breakout features mimic the expected angular geometry of breakouts.

The third example (Figure \ref{fig10}c) presents a true breakout interval in CB1 at measured depths of 171 to 173.5 m, where breakouts are developed predominantly on one side of the borehole. In this case, the lack of a well-defined opposing breakout causes the validation criteria to reject these azimuth picks, resulting in false negatives. This behavior explains the observed increase in FNR after validation, as reported in Table 1.
Taken together, these examples clarify that the breakout validation criteria act as a conservative filtering step. They significantly reduce false positives, improving the overall consistency between automatic and manual interpretations, but at the expense of discarding some asymmetric breakouts. This trade-off is explicitly reflected in the quantitative metrics. Breakout-picker achieves lower FPR and improved azimuth agreement, while exhibiting an increase in FNR after validation.

\subsubsection{Breakout width estimation}
\begin{figure*}[t]
\centering
\includegraphics[width=0.6\textwidth]{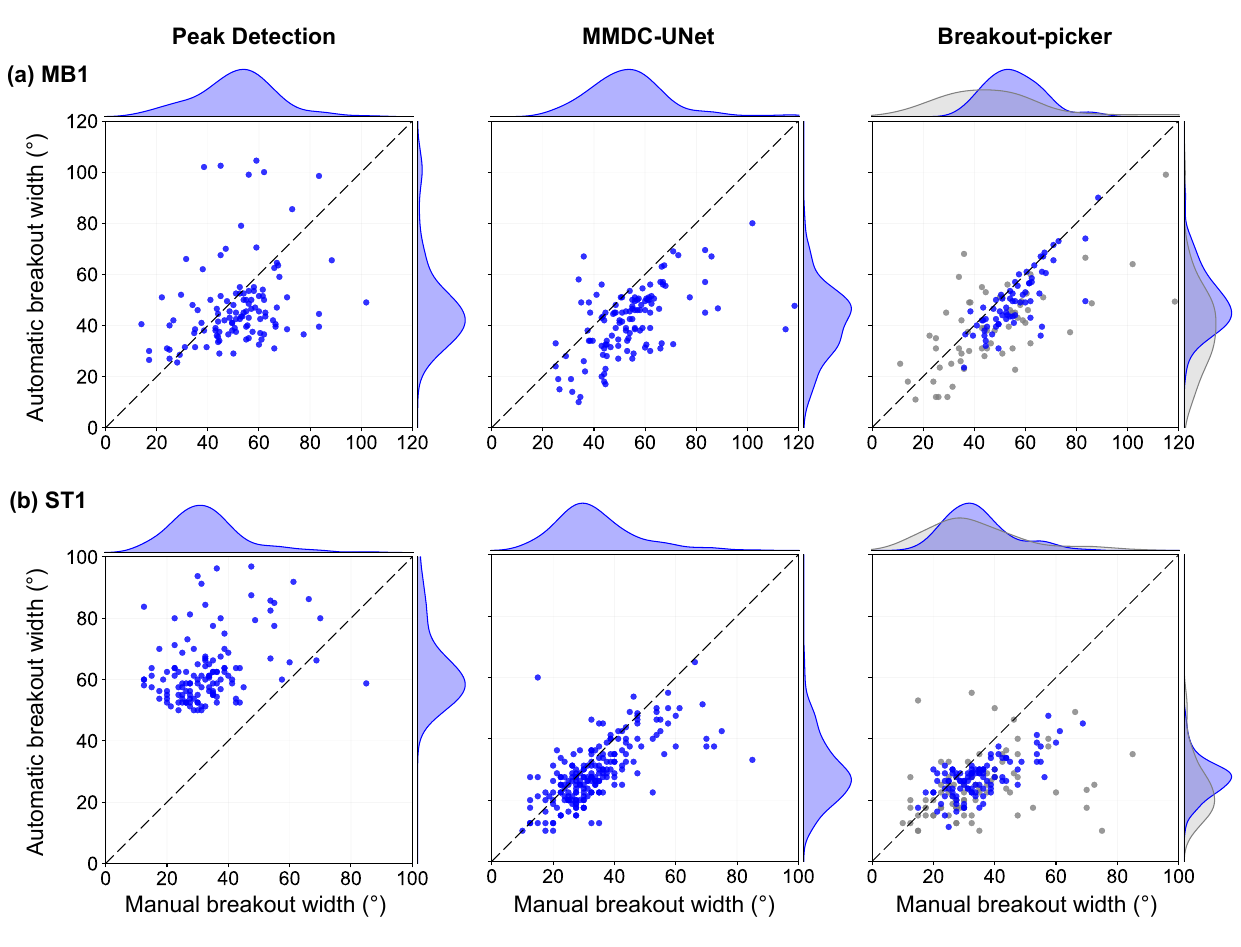}
\caption{Comparison of breakout widths estimated by different automatic methods with manual measurements for borehole CB1 and ST1. For Breakout-picker, gray points indicate breakouts removed by the validation criteria (false negatives).}
\label{fig11}
\end{figure*}

A comparison of breakout widths derived from different automatic methods with manually measured widths for CB1 and ST1 is presented in Figure \ref{fig11}. Each scatter plot shows the relationship between manual and automatic breakout widths, with the dashed line indicating perfect agreement. The manually determined breakout widths are derived from the annotated breakout regions in the acoustic image logs.

For borehole CB1, the Peak Detection method yields a width error of 14.56°, with substantial scatter and systematic deviations from the perfect match line. MMDC-UNet produces a similar width error of 14.54°. The comparable performance between these two methods in CB1 reflects the relatively high image quality and well-defined breakout boundaries in this borehole, where even simple Peak Detection can reasonably capture breakout edges. In contrast, Breakout-picker (after validation) achieves a substantially lower width error of 7.03°, with scatter points more tightly distributed close to perfect match.

For borehole ST1, the Peak Detection method produces a much larger width error of 32.54°, significantly higher than in CB1. This deterioration in performance is attributed to the poorer image quality in ST1, characterized by increased speckle noise and less clearly defined breakout boundaries. Under these challenging conditions, the segmentation-based approaches demonstrate clear advantages. MMDC-UNet achieves a width error of 7.35° and Breakout-picker (after validation) achieves 8.57°, both substantially outperforming Peak Detection. This contrast between the two boreholes highlights the robustness of segmentation-based methods in handling lower-quality acoustic image data.

The improved width estimation has direct implications for stress magnitude calculations. Breakout width is a key input parameter for constraining the horizontal stress magnitudes through analytical or numerical models. Taking ST1 as an example, the Peak Detection method yields a width error of 32.54$^\circ$, which can lead to significant overestimation of differential stress ($S_\mathrm{Hmax} - S_\mathrm{hmin}$). According to Barton et al.\cite{barton1988situ} and Vernik and Zoback \cite{vernik1992estimation}, $S_{Hmax}$ can be estimated using the following equation:
\begin{equation}
S_\mathrm{Hmax} = \frac{C_{ef} + P_f}{1 - 2\cos(\pi - 2\phi_b)} - S_\mathrm{hmin} \frac{1 + 2\cos(\pi - 2\phi_b)}{1 - 2\cos(\pi - 2\phi_b)},
\end{equation}
where $C_{ef}$ is the effective strength of the rock mass, $P_f$ is the pore pressure, and $2\phi_b$ is the breakout width. Assuming a typical deep borehole condition in diorite rock mass, with depth = 1850 m, $S_\mathrm{hmin}$ = 37 MPa, $P_f$ = 14.7 MPa (assuming hydrostatic condition), and $C_{ef}$ = 143 MPa (according to Vernik and Zoback \cite{vernik1992estimation}), a 30$^\circ$ error in breakout width estimation can result in an error of up to 17 MPa in differential stress. However, when reducing the width error to approximately 10$^\circ$, as achieved by Breakout-picker and MMDC-UNet, it lowers the estimation error in differential stress to approximately 4 MPa. Such improvement in width estimation accuracy substantially enhances the reliability of stress magnitude predictions for the retained breakout intervals, which is critical for applications such as wellbore stability analysis and hydraulic fracturing design.

\subsection{Generalization tests}
To evaluate the generalization capability of the proposed Breakout-picker beyond the BedrettoLab boreholes, we further applied the trained model to acoustic image logs acquired from two additional boreholes with data quality different from those of the training dataset. These datasets include the Hunt Well from Alberta, Canada \cite{chan2013subsurface}, and Hole 1256D drilled during the Integrated Ocean Drilling Program (IODP) in the eastern equatorial Pacific Ocean \cite{alt2007iodp}. Both boreholes exhibit abundant borehole breakouts. However, they are characterized by lower image resolution compared to the BedrettoLab data. This challenging condition make them suitable for testing the robustness and transferability of Breakout-picker.

\subsubsection{The Hunt Well}
\begin{figure*}[t]
\centering
\includegraphics[width=0.6\textwidth]{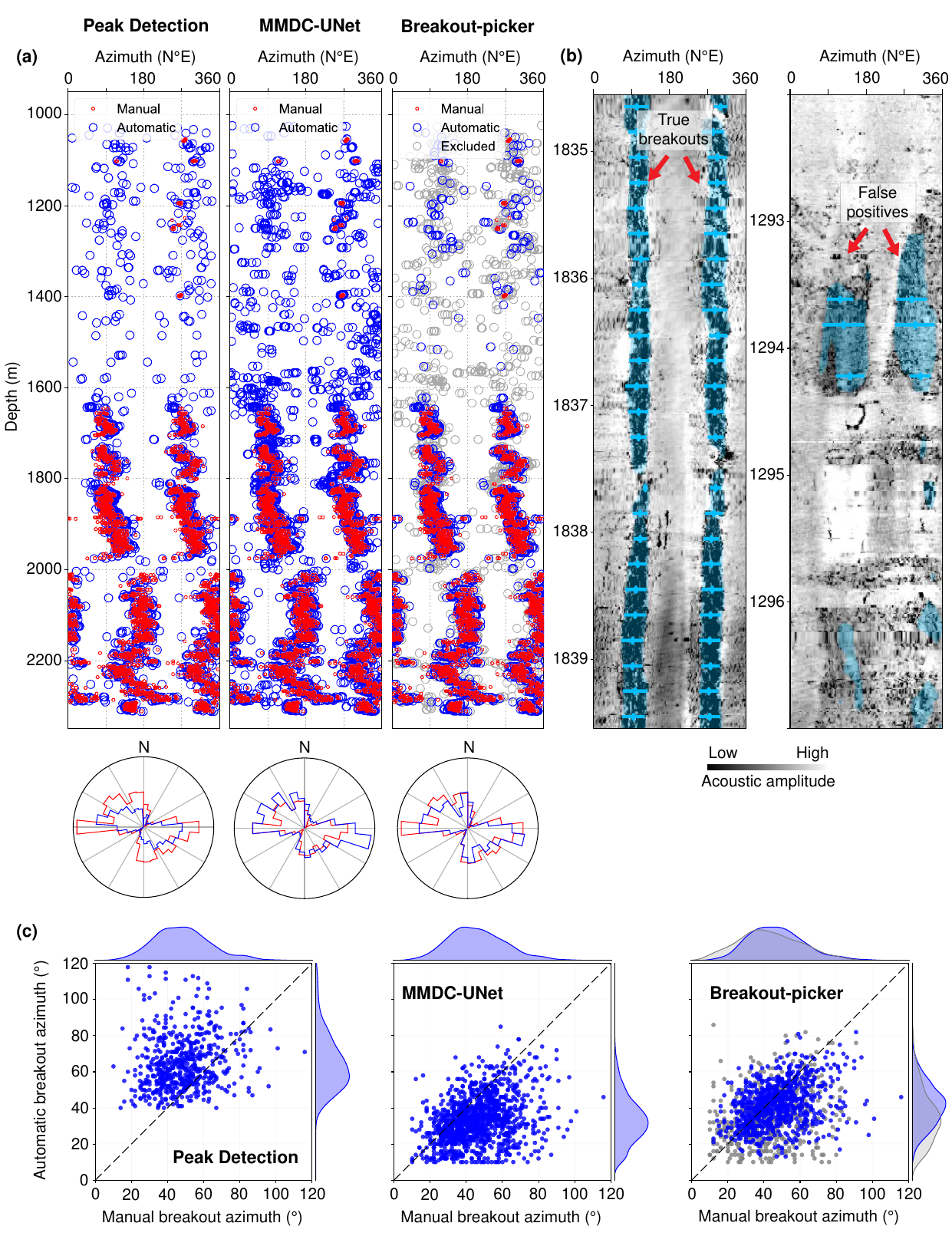}
\caption{Automatic breakout characterization results for the Hunt Well. (a) Comparison of breakout azimuths using Peak Detection, MMDC-UNet, and Breakout-picker. Red circles denote manually interpreted breakout azimuths, and blue circles denote automatically extracted azimuths. (b) Examples from a 5 m interval illustrating correct breakout detections and remaining false positives after applying the breakout validation criterion. (c) Comparison between automatically estimated breakout widths and manual measurements for the three methods.}
\label{fig12}
\end{figure*}

The Hunt Well is located in Alberta, Canada. Its total depth reaches 2363 m, penetrating 541 m of Phanerozoic sediments and 1822 m of hard crystalline metamorphic basement in the Canadian Shield \cite{chan2013subsurface}. The acoustic image log was acquired from 1006 m to 2315 m, which has been used for in situ stress characterization based on the revealed borehole breakouts \cite{wang2023heterogeneity}. Compared with the BedrettoLab boreholes, the Hunt Well acoustic image logs exhibit noticeably lower spatial resolution and reduced image contrast (Figure \ref{fig12}b), likely resulting from a fast logging speed. These factors lead to coarsely defined breakout boundaries and increased ambiguity in both amplitude and radius responses, posing challenges for automatic breakout characterization.

Figure \ref{fig12}a compares breakout azimuths derived from manual picking with those obtained using three automatic approaches: Peak Detection, MMDC-UNet, and Breakout-picker. According to the rose diagram of manual picking result, the breakouts are generally oriented in NW-SE direction, with an average azimuth of N125°E. However, significant rotations in breakout azimuth are observed between 1600 and 2400 m, causing a scattered distribution of breakout azimuth within the NW-SE quadrant. The standard deviation of breakout azimuth reaches 43°. The rose diagrams indicate that all three automatic methods can capture the primary breakout orientations. However, only the breakout azimuths determined by Breakout-picker (after validation) can correctly reflect the directional distribution.

Both the Peak Detection, MMDC-UNet, and Breakout-picker (before validation) produce highly scattered azimuth picks, particularly below 1600 m where no breakouts are identified manually. In contrast, after applying the validation criteria, the azimuths determined by Breakout-picker show closer agreement with manual picking. As summarized in Table \ref{tab1}, Breakout-picker achieves the lowest azimuth error of 5.63° among the three methods. More importantly, it reduces the FPR by 9\% and 18\% compared to the Peak Detection and MMDC-UNet methods, respectively. This improvement highlights the effectiveness of the breakout validation step in eliminating false positives. However, the validation step increases the FNR from 10\% to 46\%, reflecting its conservative nature.

Figure \ref{fig12}b shows two 5 m interval examples from the Hunt Well acoustic image log. Despite the reduced image resolution and less clearly visualized breakouts compared with the BedrettoLab data, true breakouts are correctly identified by Breakout-picker. Nevertheless, some false positives persist when the misidentified non-breakout features happen to satisfy the near-180° azimuthal symmetry criterion. These residual false positives indicate that Breakout-picker is not immune to mis-segmentation under challenging image log conditions, possibly due to the limited amount and diversity of the training samples. Breakout width comparisons are presented in Figure \ref{fig12}c. The widths estimated by Breakout-picker are in good agreement with manual measurements. It reduces the width error by approximately 9° compared to the Peak Detection method. Moreover, the width error of Breakout-picker is approximately 3° less than that of MMDC-UNet, indicating that the incorporation of negative samples does not compromise the accuracy of breakout width estimation.

\subsubsection{IODP Hole 1256D}
\begin{figure*}[t]
\centering
\includegraphics[width=0.6\textwidth]{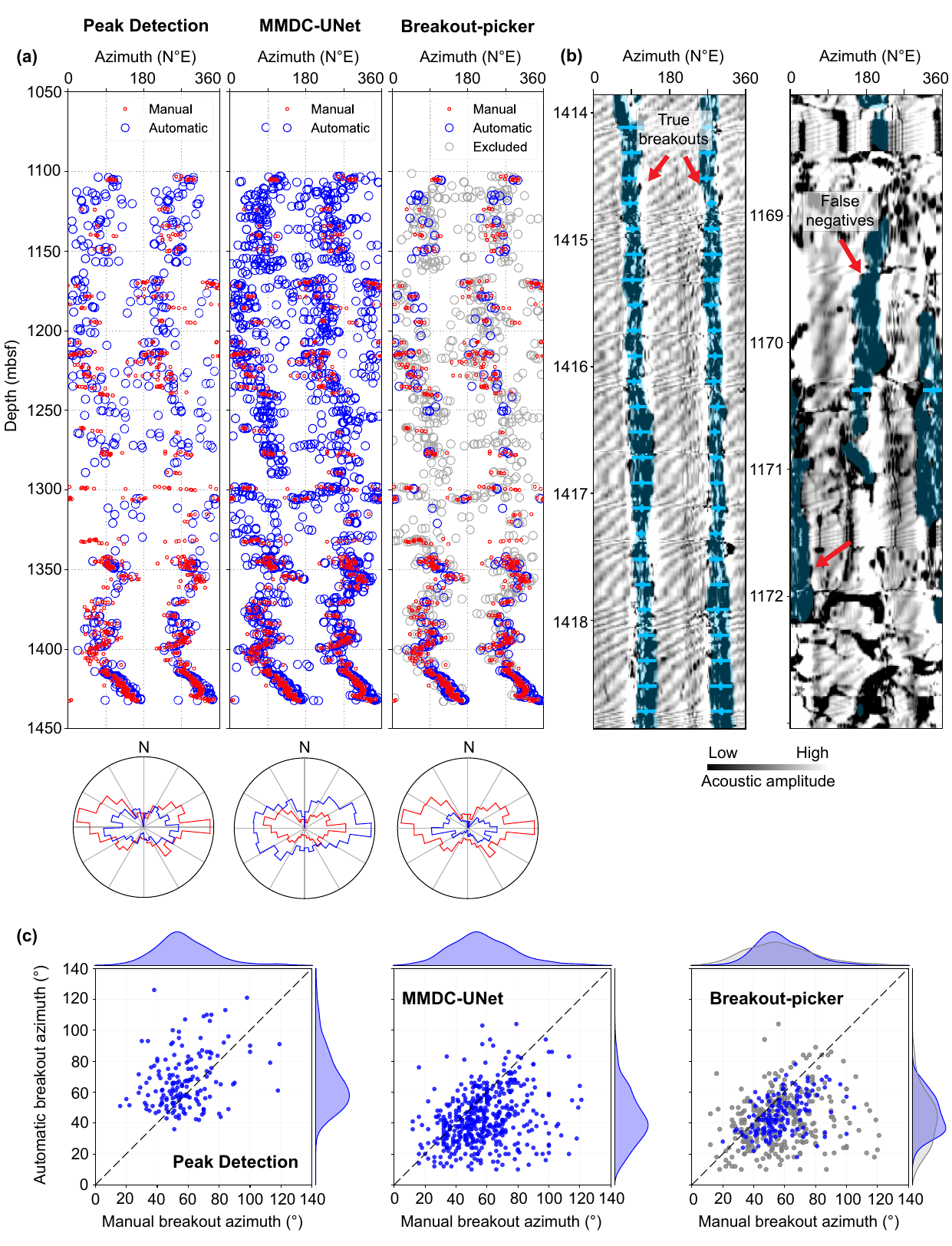}
\caption{Automatic breakout characterization results for Integrated Ocean Drilling Program (IODP) Hole 1256D. (a) Comparison of manually and automatically extracted breakout azimuths obtained using Peak Detection, MMDC-UNet, and Breakout-picker. (b) Representative depth intervals illustrating the effect of the breakout validation criterion, including retained true breakouts and excluded asymmetric or single-sided breakouts. (c) Comparison of breakout widths estimated by the three automatic methods with manual measurements.}
\label{fig13}
\end{figure*}

The second generalization test is conducted on Hole 1256D drilled during the IODP in the eastern equatorial Pacific Ocean. Hole 1256D penetrates 1507 m of basalt and sheeted dike complex in the upper oceanic crust, and encounters gabbroic rocks at 1407 m below seafloor (mbsf) \cite{wilson2006drilling, alt2007iodp}. Acoustic image logging was performed over the depth interval from 1072 to 1433 mbsf, within which abundant borehole breakouts were identified, with particular frequent appearance below approximately 1410 mbsf. Compared with the BedrettoLab dataset, the acoustic image logs from Hole 1256D exhibit lower resolution, as shown in Figure \ref{fig13}b, reflecting different logging conditions.

Figure \ref{fig13}a presents the comparison of breakout azimuths obtained using the three automatic methods and manual interpretation. The rose diagrams indicate that the breakouts in Hole 1256D are primarily oriented in a E-W trend. The directional variation of breakout is significant, resulting in a standard deviation of 42° of the breakout azimuths. As shown in the rose diagrams, breakout azimuths determined by all three automatic methods correctly capture the E-W trend of true breakout orientations. However, similar to the Hunt Well case, the Peak Detection, MMDC-UNet, and Breakout-picker before validation all produce numerous spurious detections. 

After applying the validation criteria, Breakout-picker again demonstrates improved consistency with manual interpretation. According to Table \ref{tab1}, the azimuth error for Breakout-picker is 6.59°, remaining as the lowest among the three methods. The Breakout-picker also yields the lowest FPR of 14\%, confirming the effectiveness of the proposed framework under markedly different logging conditions. However, numerous breakouts above 1350 mbsf are failed to be identified by the Breakout-picker, leading to a high FNR of 71\%. This is due to that the breakouts above 1350 mbsf are mostly asymmetric. It is reported that many basalts above 1061 mbsf are cut by sub-vertical dikes with common strongly brecciated and mineralized chilled margins, and some have doleritic textures \cite{alt2007iodp}. This lithologic heterogeneity could be the primary cause of the asymmetric breakouts. Nevertheless, the high FNR underscores the limitation of the breakout validation criteria in characterizing asymmetric breakouts.

Figure \ref{fig13}b presents zoomed-in examples illustrating both correctly identified near-symmetric breakouts and failed characterization of asymmetric breakouts. The asymmetric breakouts demonstrated in Figure \ref{fig13}b are representative of the breakout population above 1350 mbsf in Hole 1256D. Although these breakouts are correctly segmented by Breakout-picker, they are removed during the breakout validation step due to the violation of the azimuthal symmetry criterion, leading to false negative detections. The breakout width comparison shown in Figure \ref{fig13}c further indicates that, despite the exclusion of a substantial number of true breakouts during validation, the remaining identifications capture breakout widths with reasonable accuracy. This suggests that the validation step primarily affects breakout occurrence rather than the geometric characterization of retained breakout regions, even under reduced image quality.

Overall, the generalization tests on both boreholes demonstrate that Breakout-picker maintains robust performance when applied to boreholes outside the training domain, despite differences in geological setting and image resolution. Compared with the DL-based MMDC-UNet and rule-based Peak Detection approaches, Breakout-picker consistently achieves lower azimuth errors and substantially reduced FPR, while preserving accurate estimation of breakout widths. However, a notable observation emerges when examining the FPR of Breakout-picker before applying the validation step. Unlike the results for CB1 and ST1 in the BedrettoLab dataset, Breakout-picker show no significant FPR reduction compared to the MMDC-UNet in these two generalization tests (Table \ref{tab1}). 

This observation primarily reflects the limitation on the diversity of negative training samples, rather than a flaw in the negative-sample-aware training strategy. The negative samples selected from the BedrettoLab dataset may not adequately represent the hard negatives encountered in the Hunt Well and IODP Hole 1256D. Under such circumstances, the breakout validation step becomes particularly important for maintaining low FPR across diverse geological and logging settings. While the validation criteria conservatively exclude some asymmetric breakouts, this trade-off is necessary to ensure reliable breakout characterization when the model encounters previously unseen types of non-breakout features.

\section{Discussion}
\subsection{Prioritizing false-positive control in automatic breakout characterization}
A central motivation of this work is that the bottleneck in automatic breakout characterization is often not the detection of well-developed breakouts, but the suppression of false positives produced by visually similar non-breakout features in borehole images. In conventional stress interpretation workflows, breakout azimuths are treated as direct indicators of the orientation of the minimum horizontal stress \cite{wenning2017image, behboudi2023shallow}, and breakout widths are further used to infer stress magnitudes under assumptions about rock strength and borehole conditions \cite{behboudi2023shallow, schaible2025state}. In this context, false positives are particularly harmful because they can create spurious breakout intervals that bias statistical summaries of stress orientation, inflate the apparent continuity of breakout occurrence, and contaminate width distributions used for subsequent stress calculations. 

By contrast, moderate false negatives have less severe consequences for stress determination when analyzing boreholes with abundant breakout development. This is because reliable stress orientation estimates depend primarily on persistent azimuthal patterns across multiple depth intervals rather than on capturing every isolated breakout occurrence. The World Stress Map (WSM) quality-ranking system explicitly recognizes this principle through its criteria for borehole breakout data \cite{heidbach2010global}. To achieve even the minimum acceptable quality rating (C-quality, indicating $S_\mathrm{Hmax}$ orientation reliable to within $\pm$25°), the WSM requires both a minimum of four breakouts and a combined breakout length of at least 20 meters in a single well, with a standard deviation of breakout azimuths less than 25°. These requirements reflect the community agreement that stress orientation should be inferred from multiple consistent observations rather than isolated detections. Consequently, missing a subset of true breakouts (false negatives) typically does not preclude obtaining a reliable stress estimate, provided that sufficient breakouts remain to meet the WSM criteria. In contrast, even a small number of false positives can severely compromise stress interpretation by introducing spurious azimuth picks that increase the standard deviation beyond acceptable limits or create artificial breakout intervals in depth sections where no stress-induced failure actually occurred.

Given this asymmetry in how false positives and false negatives propagate through the stress analysis workflow, we argue that automatic breakout characterization should prioritize false-positive control as a first-order objective. This design principle represents a departure from conventional approaches that optimize for pixel-level segmentation agreement, which implicitly treats false positives and false negatives as equally harmful. Instead, our framework explicitly accepts a higher false negative rate in exchange for substantially reduced false positive rate, recognizing that the former can be tolerated in boreholes with abundant breakout development, whereas the latter fundamentally undermines the reliability of stress estimates.

\subsection{Role of negative-sample-aware training in reducing spurious detections}
Most segmentation-based learning pipelines implicitly assume that the background class is a single, visually heterogeneous but semantically unproblematic category. Acoustic image logs violate this assumption. The background contains structured, recurring patterns (e.g., keyseats, open fractures, tool-related artifacts) that can partially overlap with breakout signatures such as low amplitude and local radius anomalies. Under such conditions, training a network using only positive breakout examples can lead to systematic overgeneralization, because the model never learns to distinguish breakouts from visually similar non-breakout features. Our choice of using hard negative samples effectively reframes the learning objective: the network is trained not only to recognize breakouts but also to discriminate them from confounders that occupy the same feature space.

This perspective connects our approach to a broader machine learning principle. The performance in imbalanced and ambiguous detection problems often depends more on the quality and difficulty of negative examples than on network complexity. In the field of computer vision, prior works have shown that focusing training on hard negative examples can substantially improve discrimination and reduce spurious detections \cite{shrivastava2016training, xuan2020hard}. Within the breakout characterization problem, negative-sample-aware training plays a similar role by explicitly exposing the network to non-breakout structures that closely resemble breakouts. This forces the feature extractor to learn more discriminative representations between true breakouts and visually similar non-breakout features. Importantly, our results suggest that this improvement manifests primarily as a reduction in false positives, rather than a dramatic change in how true breakouts are segmented. 

However, we acknowledge that our chosen negative samples only come from a single dataset and therefore are not diverse enough to cover all the hard negative samples for breakout segmentation. Therefore, the Breakout-picker cannot avoid producing false positive segmentation results. Bigger dataset with more diverse negative samples may solve this problem.

\subsection{Trade-offs and limitations of azimuthal symmetry validation}
The breakout validation criteria adopted in this study are conservative. By enforcing a near-180° azimuthal symmetry requirement, the validation step effectively suppresses asymmetric detections that are unlikely to represent stress-induced borehole breakouts, leading to a substantial reduction in false positive rate. However, this conservative design inevitably removes true breakouts that do not satisfy the symmetry criterion, resulting in an increased false negative rate. This trade-off reflects a limitation in methodology design but acceptable for breakout characterization. As we noted, false positives in breakout characterization are generally more harmful than false negatives, as spurious breakout detections can bias stress orientation statistics and contaminate derived breakout distributions. In this context, prioritizing the suppression of false positives provides a more reliable basis for subsequent stress interpretation, even at the cost of missing a subset of true breakouts.

The results from the four borehole case studies further indicate that the breakout validation criteria do not completely eliminate false positives. In certain depth intervals, non-breakout features may still satisfy the symmetry requirement and thus remain in the final results. These residual misidentifications demonstrate that the validation criteria alone are insufficient to fully resolve all sources of ambiguity in acoustic image logs. Nevertheless, when compared with Peak Detection and MMDC-UNet, the inclusion of the validation step consistently improves overall reliability by reducing false positive detections across all test cases. This observation suggests that the proposed validation criteria, while imperfect, provide a meaningful and effective constraint that complements the CNN-based segmentation. The remaining false positives highlight opportunities for further refinement, but the current formulation already represents a clear improvement over existing automatic approaches in terms of controlling spurious breakout detections.

\section{Conclusions}
Automatic breakout characterization from acoustic image logs is often hindered by false positives, namely the misidentification of non-breakout features that resemble breakout signatures. To address this challenge, we developed Breakout-picker, a DL-based framework specifically designed to suppress false positives. It performs pixel-wise breakout segmentation using dual-channel inputs of acoustic amplitude and borehole radius logs. The network is trained with both positive breakout samples and hard negative samples that resemble breakouts. In addition, a physically motivated azimuthal symmetry validation is applied to the extracted breakout parameters to further suppress spurious detections. These designs aims at improving the reliability of automatic breakout characterization for in-situ stress analysis.

Applications to BedrettoLab acoustic image logs demonstrate that Breakout-picker achieves substantially lower false positive rates than Peak Detection and MMDC-UNet while maintaining accurate azimuth and width estimation, reflected in higher IoU scores and closer agreement with manual interpretations. Tests on the Hunt Well and IODP Hole 1256D confirm good generalization across different geological settings and logging conditions. However, limitations remain: false positives are not completely eliminated, and the conservative validation step increases false negative rates for asymmetric breakouts.

From these results, we draw two key conclusions. First, negative-sample-aware training plays a critical role in reducing spurious detections. Explicitly incorporating hard negative examples that closely resemble breakouts forces the network to learn more discriminative representations, leading to a pronounced reduction in false positives without degrading the accuracy of breakout geometry estimation. Second, post-processing constraints grounded in physical expectations, such as azimuthal symmetry, provide an effective complement to data-driven segmentation, particularly in suppressing isolated or asymmetric false detections. While such constraints inevitably introduce trade-offs, the results indicate that prioritizing false-positive control yields more reliable breakout datasets for stress interpretation, where spurious breakouts are generally more harmful than moderate omission of true breakouts. With reduced false positives, more accurate assessment of stress field heterogeneity is enabled. This is particularly critical for identifying localized stress perturbations or quantifying the wavelength of stress variation pattern, where false positives could contaminate the true stress variability.

Several directions for future improvement emerge from this study. Increasing the diversity of negative samples across different geological settings and logging conditions may further enhance the generalization ability over various datasets. In addition, replacing the current binary symmetry criterion with a more flexible formulation and incorporating uncertainty information could help better balance false positive and false negative control. Overall, this study demonstrates that combining negative-sample-aware deep learning with physically motivated validation provides a promising pathway toward more reliable and transferable automatic breakout characterization from acoustic image logs.

\section*{Data availability}
The acoustic image logs from the BedrettoLab boreholes are available on ETH Research Collection \cite{broker2023borehole}: https://doi.org/10.3929/ethz-b-000616727. The acoustic image log from the Hunt Well is available on the Canadian Dataverse Repository \cite{schmitt2022geophysical}: https://doi.org/10.5683/SP3/YYNVW8. The acoustic image log from the IODP Hole 1256D is available on the IODP data repository: https://mlp.ldeo.columbia.edu/data/iodp-usio/exp312/1256D/.  In addition, we release the annotated breakout dataset \cite{wang2026acoustic}(https:/doi.org/10.5281/zenodo.18627151) and the source codes of Breakout-picker (https://github.com/wanggy-1/Breakout-picker/tree/main) to facilitate future research.

\bibliographystyle{IEEEtran}
\bibliography{BreakoutPicker-refs}

@article{alt2007iodp,
   author = {Alt, J. C. and Teagle, D. A. H. and Umino, S. and Miyashita, S. and Banerjee, N. R. and Wilson, D. S. and the, Iodp Expeditions and Scientists and the, O. D. P. Leg Scientific Party},
   title = {IODP Expeditions 309 and 312 Drill an Intact Section of Upper Oceanic Basement into Gabbros},
   journal = {Sci. Dril.},
   volume = {4},
   pages = {4-10},
   ISSN = {1816-3459},
   DOI = {10.2204/iodp.sd.4.01.2007},
   year = {2007},
   type = {Journal Article}
}

@article{barton1988situ,
   author = {Barton, Colleen A. and Zoback, Mark D. and Burns, Kerry L.},
   title = {In-situ stress orientation and magnitude at the Fenton Geothermal Site, New Mexico, determined from wellbore breakouts},
   journal = {Geophysical Research Letters},
   volume = {15},
   number = {5},
   pages = {467-470},
   ISSN = {0094-8276},
   DOI = {/10.1029/GL015i005p00467},
   year = {1988},
   type = {Journal Article}
}

@article{bashmagh2024application,
   author = {Bashmagh, Nazir and Lin, Weiren and Ishitsuka, Kazuya},
   title = {Application of supervised machine learning classification models to identify borehole breakouts in carbonate reservoirs based on conventional log data},
   journal = {International Journal of the JSRM},
   volume = {20},
   number = {1},
   DOI = {10.11187/ijjsrm.240101},
   year = {2024},
   type = {Journal Article}
}

@article{behboudi2023shallow,
   author = {Behboudi, E. and McNamara, D. D. and Lokmer, I.},
   title = {Shallow Tectonic Stress Magnitudes at the Hikurangi Subduction Margin, New Zealand},
   journal = {Geochemistry, Geophysics, Geosystems},
   volume = {24},
   number = {10},
   pages = {e2022GC010836},
   DOI = {10.1029/2022GC010836},
   year = {2023},
   type = {Journal Article}
}

@article{bell1979northeast,
   author = {Bell, J. S. and Gough, D. I.},
   title = {Northeast-southwest compressive stress in Alberta evidence from oil wells},
   journal = {Earth and Planetary Science Letters},
   volume = {45},
   number = {2},
   pages = {475-482},
   ISSN = {0012-821X},
   DOI = {10.1016/0012-821X(79)90146-8},
   year = {1979},
   type = {Journal Article}
}

@inproceedings{blake2012borehole,
   author = {Blake, Kelly and Davatzes, Nicholas C},
   title = {Borehole image log and statistical analysis of FOH-3D, Fallon Naval Air Station, NV},
   booktitle = {Proceedings of the thirty-seventh workshop on geothermal reservoir engineering},
   year={2012}, 
   type = {Conference Proceedings}
}

@article{blumling1983orientation,
   author = {Blümling, P. and Fuchs, K. and Schneider, T.},
   title = {Orientation of the stress field from breakouts in a crystalline well in a seismic active area},
   journal = {Physics of the Earth and Planetary Interiors},
   volume = {33},
   number = {4},
   pages = {250-254},
   ISSN = {0031-9201},
   DOI = {10.1016/0031-9201(83)90042-0},
   year = {1983},
   type = {Journal Article}
}

@phdthesis{chan2013subsurface,
   author = {Chan, Judith S},
   title = {Subsurface geophysical characterization of the crystalline Canadian Shield in Northeastern Alberta: implications for geothermal development},
   DOI = {10.7939/R3BR8MQ6B},
   year = {2013},
   school = {University of Alberta}
}

@inproceedings{chen2018encoder,
   author = {Chen, Liang-Chieh and Zhu, Yukun and Papandreou, George and Schroff, Florian and Adam, Hartwig},
   title = {Encoder-decoder with atrous separable convolution for semantic image segmentation},
   booktitle = {Proceedings of the European conference on computer vision (ECCV)},
   pages = {801-818}, 
   year = {2018}, 
   DOI = {10.1007/978-3-030-01234-2_49},
   type = {Conference Proceedings}
}

@article{cunha2025semi,
   author = {Cunha, A.Í. Da and Anjos, G. Brandemburg Dos and Nascimento, R. and de Jesus, C. Menezes and Jauregui, J.J.H.},
   title = {A Semi-Supervised Approach for Borehole Image Log Structures Segmentation for Small Labeled Datasets},
   volume = {2025},
   number = {1},
   pages = {1-5},
   ISSN = {2214-4609},
   DOI = {10.3997/2214-4609.2025640028},
   year = {2025},
   type = {Journal Article}
}

@article{dias2020automatic,
   author = {Dias, Luciana Olivia and Bom, Clécio R and Faria, Elisangela L and Valentín, Manuel Blanco and Correia, Maury Duarte and de Albuquerque, Márcio P and de Albuquerque, Marcelo P and Coelho, Juliana M},
   title = {Automatic detection of fractures and breakouts patterns in acoustic borehole image logs using fast-region convolutional neural networks},
   journal = {Journal of Petroleum Science and Engineering},
   volume = {191},
   pages = {107099},
   ISSN = {0920-4105},
   DOI = {10.1016/j.petrol.2020.107099},
   year = {2020},
   type = {Journal Article}
}

@article{duan2016evolution,
   author = {Duan, K. and Kwok, C. Y.},
   title = {Evolution of stress-induced borehole breakout in inherently anisotropic rock: Insights from discrete element modeling},
   journal = {Journal of Geophysical Research: Solid Earth},
   volume = {121},
   number = {4},
   pages = {2361-2381},
   ISSN = {2169-9313},
   DOI = {10.1002/2015JB012676},
   year = {2016},
   type = {Journal Article}
}

@article{haimson2007micromechanisms,
   author = {Haimson, B.},
   title = {Micromechanisms of borehole instability leading to breakouts in rocks},
   journal = {International Journal of Rock Mechanics and Mining Sciences},
   volume = {44},
   number = {2},
   pages = {157-173},
   ISSN = {1365-1609},
   DOI = {10.1016/j.ijrmms.2006.06.002},
   year = {2007},
   type = {Journal Article}
}

@article{heidbach2010global,
   author = {Heidbach, Oliver and Tingay, Mark and Barth, Andreas and Reinecker, John and Kurfeß, Daniel and Müller, Birgit},
   title = {Global crustal stress pattern based on the World Stress Map database release 2008},
   journal = {Tectonophysics},
   volume = {482},
   number = {1},
   pages = {3-15},
   ISSN = {0040-1951},
   DOI = {10.1016/j.tecto.2009.07.023},
   year = {2010},
   type = {Journal Article}
}

@article{ito1998stress,
   author = {Ito, T. and Kurosawa, K. and Hayashi, K.},
   title = {Stress Concentration at the Bottom of a Borehole and its Effect on Borehole Breakout Formation},
   journal = {Rock Mechanics and Rock Engineering},
   volume = {31},
   number = {3},
   pages = {153-168},
   ISSN = {1434-453X},
   DOI = {10.1007/s006030050016},
   year = {1998},
   type = {Journal Article}
}

@article{lin2010localized,
   author = {Lin, Weiren and Yeh, En-Chao and Hung, Jih-Hao and Haimson, Bezalel and Hirono, Tetsuro},
   title = {Localized rotation of principal stress around faults and fractures determined from borehole breakouts in hole B of the Taiwan Chelungpu-fault Drilling Project (TCDP)},
   journal = {Tectonophysics},
   volume = {482},
   number = {1-4},
   pages = {82-91},
   ISSN = {0040-1951},
   DOI = {10.1016/j.tecto.2009.06.020},
   year = {2010},
   type = {Journal Article}
}

@article{ma2022multi,
   author = {Ma, X. and Hertrich, M. and Amann, F. and Bröker, K. and Gholizadeh Doonechaly, N. and Gischig, V. and Hochreutener, R. and Kästli, P. and Krietsch, H. and Marti, M. and Nägeli, B. and Nejati, M. and Obermann, A. and Plenkers, K. and Rinaldi, A. P. and Shakas, A. and Villiger, L. and Wenning, Q. and Zappone, A. and Bethmann, F. and Castilla, R. and Seberto, F. and Meier, P. and Driesner, T. and Loew, S. and Maurer, H. and Saar, M. O. and Wiemer, S. and Giardini, D.},
   title = {Multi-disciplinary characterizations of the BedrettoLab – a new underground geoscience research facility},
   journal = {Solid Earth},
   volume = {13},
   number = {2},
   pages = {301-322},
   ISSN = {1869-9529},
   DOI = {10.5194/se-13-301-2022},
   year = {2022},
   type = {Journal Article}
}

@article{moos1990utilization,
   author = {Moos, Daniel and Zoback, Mark D},
   title = {Utilization of observations of well bore failure to constrain the orientation and magnitude of crustal stresses: application to continental, Deep Sea Drilling Project, and Ocean Drilling Program boreholes},
   journal = {Journal of Geophysical Research: Solid Earth},
   volume = {95},
   number = {B6},
   pages = {9305-9325},
   ISSN = {0148-0227},
   DOI = {10.1029/JB095iB06p09305},
   year = {1990},
   type = {Journal Article}
}

@article{plumb1985stress,
   author = {Plumb, Richard and Hickman, Stephen},
   title = {Stress-induced borehole elongation: A comparison between the four-arm dipmeter and the borehole televiewer in the Auburn Geothermal Well},
   journal = {Journal of Geophysical Research},
   volume = {90},
   pages = {5513-5521},
   DOI = {10.1029/JB090iB07p05513},
   year = {1985},
   type = {Journal Article}
}

@article{rajabi2010subsurface,
   author = {Rajabi, Mojtaba and Sherkati, Shahram and Bohloli, Bahman and Tingay, Mark},
   title = {Subsurface fracture analysis and determination of in-situ stress direction using FMI logs: An example from the Santonian carbonates (Ilam Formation) in the Abadan Plain, Iran},
   journal = {Tectonophysics},
   volume = {492},
   number = {1},
   pages = {192-200},
   ISSN = {0040-1951},
   DOI = {10.1016/j.tecto.2010.06.014},
   year = {2010},
   type = {Journal Article}
}

@article{sahara2014impact,
   author = {Sahara, David P and Schoenball, Martin and Kohl, Thomas and Müller, Birgit IR},
   title = {Impact of fracture networks on borehole breakout heterogeneities in crystalline rock},
   journal = {International Journal of Rock Mechanics and Mining Sciences},
   volume = {71},
   pages = {301-309},
   ISSN = {1365-1609},
   DOI = {10.1016/j.ijrmms.2014.07.001},
   year = {2014},
   type = {Journal Article}
}

@article{schaible2025state,
   author = {Schaible, Kaitlin E. and Saffer, Demian M.},
   title = {State of Stress Across Major Faults in the Nankai Subduction Zone Estimated From Wellbore Breakouts},
   journal = {Journal of Geophysical Research: Solid Earth},
   volume = {130},
   number = {7},
   pages = {e2024JB030242},
   ISSN = {2169-9313},
   DOI = {10.1029/2024JB030242},
   year = {2025},
   type = {Journal Article}
}

@inproceedings{shrivastava2016training,
   author = {Shrivastava, Abhinav and Gupta, Abhinav and Girshick, Ross},
   title = {Training region-based object detectors with online hard example mining},
   booktitle = {Proceedings of the IEEE conference on computer vision and pattern recognition},
   pages = {761-769},
   DOI = {10.1109/cvpr.2016.89 },
   type = {Conference Proceedings}
}

@article{valley2019stress,
   author = {Valley, Benoît and Evans, Keith F.},
   title = {Stress magnitudes in the Basel enhanced geothermal system},
   journal = {International Journal of Rock Mechanics and Mining Sciences},
   volume = {118},
   pages = {1-20},
   ISSN = {1365-1609},
   DOI = {10.1016/j.ijrmms.2019.03.008},
   year = {2019},
   type = {Journal Article}
}

@article{vernik1992estimation,
   author = {Vernik, Lev and Zoback, Mark D.},
   title = {Estimation of maximum horizontal principal stress magnitude from stress-induced well bore breakouts in the Cajon Pass Scientific Research borehole},
   journal = {Journal of Geophysical Research: Solid Earth},
   volume = {97},
   number = {B4},
   pages = {5109-5119},
   ISSN = {0148-0227},
   DOI = {https://doi.org/10.1029/91JB01673},
   year = {1992},
   type = {Journal Article}
}

@inproceedings{wang2020automated,
   author = {Wang, W and Schmitt, DR},
   title = {Automated borehole breakout interpretation from ultrasonic imaging: Application to a deep borehole drilled into the crystalline crust},
   booktitle = {ARMA US Rock Mechanics/Geomechanics Symposium},
   publisher = {ARMA},
   pages = {ARMA-2020-1270},
   ISBN = {0979497558},
   type = {Conference Proceedings}
}

@article{wang2023heterogeneity,
   author = {Wang, Wenjing and Schmitt, Douglas R and Chan, Judith},
   title = {Heterogeneity versus anisotropy and the state of stress in stable cratons: Observations from a deep borehole in Northeastern Alberta, Canada},
   journal = {Journal of Geophysical Research: Solid Earth},
   volume = {128},
   number = {3},
   pages = {e2022JB025287},
   ISSN = {2169-9313},
   DOI = {10.1029/2022JB025287},
   year = {2023},
   type = {Journal Article}
}

@article{wenning2017image,
   author = {Wenning, Quinn C. and Berthet, Theo and Ask, Maria and Zappone, Alba and Rosberg, Jan-Erik and Almqvist, Bjarne S. G.},
   title = {Image log analysis of in situ stress orientation, breakout growth, and natural geologic structures to 2.5 km depth in central Scandinavian Caledonides: Results from the COSC-1 borehole},
   journal = {Journal of Geophysical Research: Solid Earth},
   volume = {122},
   number = {5},
   pages = {3999-4019},
   ISSN = {2169-9313},
   DOI = {10.1002/2016JB013776},
   year = {2017},
   type = {Journal Article}
}

@inproceedings{wessling2011automated,
   author = {Wessling, Stefan and Dahl, Thomas and Pantic, Dinah and Pei, Jianyong},
   title = {Automated Breakout Detection Current Possibilities And Their Benefit For Wellbore Stability Prediction},
   booktitle = {SPWLA 52nd Annual Logging Symposium},
   address = {SPWLA-2011-T},
   volume = {All Days},
   type = {Conference Proceedings}
}

@article{wilson2006drilling,
   author = {Wilson, Douglas S. and Teagle, Damon A. H. and Alt, Jeffrey C. and Banerjee, Neil R. and Umino, Susumu and Miyashita, Sumio and Acton, Gary D. and Anma, Ryo and Barr, Samantha R. and Belghoul, Akram and Carlut, Julie and Christie, David M. and Coggon, Rosalind M. and Cooper, Kari M. and Cordier, Carole and Crispini, Laura and Durand, Sedelia Rodriguez and Einaudi, Florence and Galli, Laura and Gao, Yongjun and Geldmacher, Jörg and Gilbert, Lisa A. and Hayman, Nicholas W. and Herrero-Bervera, Emilio and Hirano, Nobuo and Holter, Sara and Ingle, Stephanie and Jiang, Shijun and Kalberkamp, Ulrich and Kerneklian, Marcie and Koepke, Jürgen and Laverne, Christine and Vasquez, Haroldo L. Lledo and Maclennan, John and Morgan, Sally and Neo, Natsuki and Nichols, Holly J. and Park, Sung-Hyun and Reichow, Marc K. and Sakuyama, Tetsuya and Sano, Takashi and Sandwell, Rachel and Scheibner, Birgit and Smith-Duque, Chris E. and Swift, Stephen A. and Tartarotti, Paola and Tikku, Anahita A. and Tominaga, Masako and Veloso, Eugenio A. and Yamasaki, Toru and Yamazaki, Shusaku and Ziegler, Christa},
   title = {Drilling to Gabbro in Intact Ocean Crust},
   journal = {Science},
   volume = {312},
   number = {5776},
   pages = {1016-1020},
   DOI = {10.1126/science.1126090},
   year = {2006},
   type = {Journal Article}
}

@inproceedings{xuan2020hard,
   author = {Xuan, Hong and Stylianou, Abby and Liu, Xiaotong and Pless, Robert},
   title = {Hard Negative Examples are Hard, but Useful},
   series = {Computer Vision – ECCV 2020},
   publisher = {Springer International Publishing},
   pages = {126-142},
   ISBN = {978-3-030-58568-6},
   DOI = {10.1007/978-3-030-58568-6_8},
   type = {Conference Proceedings}
}

@inproceedings{yang2022automated,
   author = {Yang, J and Goodfellow, SD and Harrison, JP},
   title = {Automated extraction of borehole breakout properties from acoustic televiewer (ATV) data},
   booktitle = {ARMA US Rock Mechanics/Geomechanics Symposium},
   publisher = {ARMA},
   pages = {ARMA-2022-0408},
   ISBN = {0979497574},
   DOI = {10.56952/ARMA-2022-0408},
   type = {Conference Proceedings}
}

@article{yeom2025automatic,
   author = {Yeom, Juyeol and Kim, Hayoung and Chang, Chandong and Jo, Yeonguk and Chen, Zhuoheng and Lee, Kyungbook},
   title = {Automatic detection of borehole breakout for image logs using YOLO algorithm},
   journal = {Geoenergy Science and Engineering},
   volume = {252},
   pages = {213925},
   ISSN = {2949-8910},
   DOI = {10.1016/j.geoen.2025.213925},
   year = {2025},
   type = {Journal Article}
}

@article{zhang2023fault,
   author = {Zhang, Shihuai and Ma, Xiaodong and Bröker, Kai and van Limborgh, Rutger and Wenning, Quinn and Hertrich, Marian and Giardini, Domenico},
   title = {Fault zone spatial stress variations in a granitic rock mass: Revealed by breakouts within an array of boreholes},
   journal = {Journal of Geophysical Research: Solid Earth},
   volume = {128},
   number = {8},
   pages = {e2023JB026477},
   ISSN = {2169-9313},
   DOI = {10.1029/2023JB026477},
   year = {2023},
   type = {Journal Article}
}

@article{zoback2003determination,
   author = {Zoback, Mark D and Barton, CA and Brudy, M and Castillo, DA and Finkbeiner, Thomas and Grollimund, BR and Moos, DB and Peska, Pl and Ward, CD and Wiprut, DJ},
   title = {Determination of stress orientation and magnitude in deep wells},
   journal = {International Journal of Rock Mechanics and Mining Sciences},
   volume = {40},
   number = {7-8},
   pages = {1049-1076},
   ISSN = {1365-1609},
   DOI = {10.1016/j.ijrmms.2003.07.001},
   year = {2003},
   type = {Journal Article}
}

@article{zoback1985well,
   author = {Zoback, Mark D and Moos, Daniel and Mastin, Larry and Anderson, Roger N},
   title = {Well bore breakouts and in situ stress},
   journal = {Journal of Geophysical Research: Solid Earth},
   volume = {90},
   number = {B7},
   pages = {5523-5530},
   ISSN = {0148-0227},
   DOI = {10.1029/JB090iB07p05523},
   year = {1985},
   type = {Journal Article}
}

@inproceedings{thomas2015structural,
  title={Structural interpretation from Televiewer surveys},
  author={Thomas, RDH and Neilsen, JM and Wilson, HF and Lamb, P},
  booktitle={FMGM 2015: Proceedings of the Ninth Symposium on Field Measurements in Geomechanics},
  pages={729--741},
  year={2015},
  organization={Australian Centre for Geomechanics}, 
  doi={10.36487/ACG_rep/1508_53_Thomas}
}

@dataset{broker2023borehole,
  title={Borehole logging data (acoustic and optical televiewer) from boreholes in the Bedretto Lab [Dataset]},
  author={Br{\"o}ker, Kai Erich Norbert and Wenning, Quinn and Hertrich, Marian and Ma, Xiaodong},
  year={2023},
  publisher={ETH Zurich}, 
  doi={10.3929/ethz-b-000616727}, 
  url={https://doi.org/10.3929/ethz-b-000616727}
}

@dataset{schmitt2022geophysical,
  title={Geophysical logging and image Data from the Hunt Well, NE Alberta [Dataset]},
  author={Schmitt, DR and Wang, W and Chan, J},
  year={2022}, 
  publisher={Borealis}, 
  doi={10.5683/SP3/YYNVW8}, 
  url={https://doi.org/10.5683/SP3/YYNVW8}
}

@dataset{wang2026acoustic,
  author       = {Wang, Guangyu},
  title        = {Acoustic image log borehole breakout dataset for
                   advancing deep learning in automatic breakout
                   characterization from image logs
                  },
  month        = feb,
  year         = 2026,
  publisher    = {Zenodo},
  doi          = {10.5281/zenodo.18627151},
  url          = {https://doi.org/10.5281/zenodo.18627151},
}

\vfill

\end{document}